\documentclass[11pt]{article}
\pdfoutput=1

\usepackage{amsmath,amsthm,verbatim,amssymb,amsfonts,amscd, graphicx}
\usepackage{graphics}
\usepackage{centernot}
\theoremstyle{plain}
\newtheorem{theorem}{Theorem}

\newtheorem{proposition}{Proposition}

\theoremstyle{definition}

\usepackage{wrapfig}

\newcommand{\laprop}{\textsc{LaProp}}
\newcommand{\sgd}{\textsc{SGD}}
\newcommand{\adam}{\textsc{Adam}}
\newcommand{\amsgrad}{\textsc{AMSGrad}}

\newcommand{\rmsprop}{\textsc{RMSProp}}

\usepackage[font=small]{caption}
\usepackage{subcaption}
\usepackage{helvet}  %Required
\usepackage{courier}  %Required
\usepackage{url}  %Required
\usepackage{graphicx}  %Required
\usepackage{multirow}
\usepackage{amsthm}
\usepackage{color}
\usepackage{MnSymbol}
\usepackage{makecell}
\usepackage{arydshln}
\usepackage{amsmath}
\usepackage{xcolor}
\usepackage{caption} 
\usepackage[numbers,sort&compress]{natbib}
\usepackage{textcomp}
\usepackage{wrapfig}
\usepackage{algorithm}
\usepackage{algorithmic}

\usepackage[in,plain]{fullpage}
\newcommand{\footremember}[2]{%
    \footnote{#2}
    \newcounter{#1}
    \setcounter{#1}{\value{footnote}}%
}
\newcommand{\footrecall}[1]{%
    \footnotemark[\value{#1}]%
} 

\begin{document}
%\[ \fbox{$\Box$} \fbox{$\hat\Box$} \fbox{$\tilde\Box$} \]

\title{LaProp: Separating Momentum and Adaptivity in Adam}%:\\ %\textit{Optimal Label Smoothing}}
\author{Liu Ziyin\footremember{1}{Department of Physics and Institute for Physics of Intelligence, University of Tokyo, 7-3-1 Hongo, Bunkyo-ku, Tokyo 113-0033, Japan}\footnote{Correspondence
to: Liu Ziyin <zliu@cat.phys.s.u-tokyo.ac.jp>} \and Zhikang T. Wang\footrecall{1} \and Masahito Ueda\footrecall{1} \footnote{RIKEN Center for Emergent Matter Science (CEMS), Wako, Saitama 351-0198, Japan} }
%\email{zliu@cat.phys.s.u-tokyo.ac.jp}

\date{}
\maketitle

\begin{abstract}
     We identity a by-far-unrecognized problem of \adam-style optimizers which results from unnecessary coupling between momentum and adaptivity. The coupling leads to instability and divergence when the momentum and adaptivity parameters are mismatched. In this work, we propose a method, \laprop, which decouples momentum and adaptivity in the \adam-style methods. We show that the decoupling leads to greater flexibility in the hyperparameters and allows for a straightforward interpolation between the signed gradient methods and the adaptive gradient methods. We experimentally show that \laprop\ has consistently improved speed and stability over \adam\ on a variety of tasks. We also bound the regret of \laprop\ on a convex problem and show that our bound differs from that of \adam\ by a key factor, which demonstrates its advantage. %We experimentally show that \laprop\ outperforms the previous methods on a toy task with noisy gradients, optimization of extremely deep fully-connected networks, neural art style transfer, natural language processing using transformers, and reinforcement learning with deep-Q networks. The performance improvement of \laprop\ is shown to be consistent, sometimes dramatic and qualitative. 
    
    %\laprop\ eliminates the need for gradient clipping, and reduces the number of hyperparameter by two (compared to Adam). It is shown to outperform other existing methods (e.g. Adam, RMSProp, amsgrad etc.) of similar computational complexity by a large margin, and is capable of optimizing hard and unstable problems smoothly. In particular, we demonstrate that \laprop\ can optimize a 5000 layer fully connected ReLu network on MNIST, without special hyperparameter tuning or gradient clipping, while other methods either diverge or does not start training. We also demonstrate that our method also outperform existing methods on large scale datasets and modern architectures and in reinforcement learning.
    %In this work, we use concepts from Langevin dynamics to provide insights for subgradient methods such as \rmsprop\ and \adam\ works. However, our finding suggests a drastic change to Adam, making it more like \rmsprop\ with Momentum. However, we show that the advantage of this new optimizer is huge. It is both qualitatively different from \adam\ and \rmsprop. In particular, it is able to optimize upto $45$-depth linear and non-linear networks with almost no decrease in test performance or training speed. Due to our motivation from Langevin dynamics, we call our new optimizer \laprop.
\end{abstract}

\vspace{-3mm}
\section{Introduction}
\vspace{-2mm}
Modern deep learning research and application have become increasingly time-consuming due to the need to train and evaluate large models on large-scale problems, where the training can take weeks or even months to complete. This makes the study of optimization of neural networks an important field of research \cite{ruder2016overview, sun2019optimization}.  At the core of the research lies the optimizers, i.e.~the algorithms by which the parameters of neural networks are updated. Since the optimizer \adam\ was proposed and became widely used, various modifications of \adam\ have also been proposed to overcome the difficulties encountered in specific cases \cite{Reddi2018convergence, liu2019variance, loshchilov2017fixing, Luo2019AdaBound, chen2018closing}. Nevertheless, none of them have shown consistent improvement over \adam\ on all tasks without using additional hyperparameters, and the mechanism behind remains vague. 

In this paper, we propose a new adaptive optimizer, \laprop. When compared with \adam\ on a variety of tasks, \laprop\ consistently performs better or at least comparably, and especially, we find that \laprop\ performs better when the task is noisy or unstable, and it can optimize difficult problems for which \adam\ fails. Such improvement comes almost for free: \laprop\ is closely related to \adam, and it has exactly the same number of hyperparameters as \adam; the hyperparameter settings of \adam\ can be readily carried over to \laprop. Moreover, \laprop\ is more stable and it allows for a wider range of hyperparameter choice for which \adam\ would diverge, which also makes \laprop\ possible to reach a higher performance over it wider hyperparameter range. We hope that our proposed optimizer benefits future study of deep learning and industrial applications in general.

This work has three contributions: (1) proposing a new adaptive optimization algorithm that has considerably better stability and flexibility, and confirming that such advantages indeed translate to wider applicability and better performance on tasks that are relevant to industrial applications and academic interests; (2) conceptually, extending an existing framework for understanding different optimizers, as none of the previously proposed frameworks include our method as a subclass; (3) theoretically, we give a convergence proof of our method and show that our method is different from the previous ones, i.e.~\adam\ and \amsgrad, by a key factor that has limited their flexibility and may lead to worse performance.
\begin{wrapfigure}[10]{R}{180pt}
\vspace{-7mm}
    \centering
    \includegraphics[width=150pt]{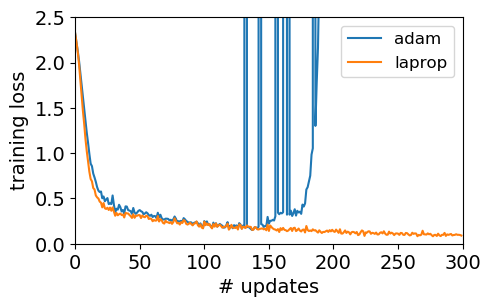}
    \caption{\label{fig:adam-divergence}Divergence of \adam\ on a two-layer ReLU network trained on MNIST with $\mu=0.9,\ \nu=0.7$. In contrast, \laprop\ is always stable.}
\end{wrapfigure}

In section~\ref{sec: framework}, we extend the framework in Ref.~\cite{wilson2017marginal} to give an overview of the up-to-date optimization algorithms and introduce \laprop\ as an alternative to \adam. In section~\ref{sec: interpolation}, we show that \laprop\ is flexible and can be tuned to interpolate between the adaptive and the signed gradient methods, while it is impossible for \adam, and in section~\ref{sec: theory}, we give some mathematical properties of our method and bound its regret by $O(\sqrt{T})$. The experiments are presented in section~\ref{sec: experiment}.
\vspace{-4mm}
\section{Conceptual Framework}\label{sec: framework}
\vspace{-2mm}
Many attempts at presenting all adaptive optimizers within the same framework exist, such as in Ref.~\cite{Reddi2018convergence, Luo2019AdaBound, chen2018convergence}. However, likely due to the popularization of \adam, all those frameworks assume the structure of the \adam\ algorithm, where a momentum $m_t$ and a preconditioner $n_t$ are computed from the gradient separately and then combined to give an update $m_t/\sqrt{n_t}$, and as a result, those frameworks do not include our adaptive optimizer as a subclass. Below, we generalize the framework in Ref.~\cite{wilson2017marginal} to include our optimizer \laprop, and we review the up-to-date optimization algorithms and discuss the relation between \laprop\ and them under the generalized framework.
\vspace{-4mm}
\subsection{Gradient Descent}
\vspace{-2mm}
Let $\ell(\theta)$ be the loss function we want to minimize, and $\theta$ the model parameters that we optimize. To minimize $\ell$, stochastic gradient descent (SGD) performs an update step $\theta_{t+1} = \theta_t - \lambda_t \nabla \ell(\theta_{t})$, where $\lambda$ is a step size. The optimization can be sped up by introducing momentum, which uses past update steps to accelerate the current step. The update rule can be expressed as
\begin{equation}
    \theta_{t+1} = \theta_t - \lambda_t \nabla \ell[\theta_{t} - \gamma_t(\theta_t - \theta_{t-1})] + \mu_t(\theta_t - \theta_{t-1}),\qquad 0\leq\mu_t<1,
\end{equation}
where $\mu_t$ is the momentum hyperparameter controlling how the momentum decays. Here the factor $1-\mu_t$ before $\lambda_t$ is neglected for clarity. The Heavy-Ball momentum has $\gamma_t=0$, and the Nesterov momentum has $\gamma_t = \mu_t$. We assume $\gamma_t=0$ in the following discussion. Momentum has been proven to accelerate convergence \cite{nesterov1983method}, and it is an experimental fact that optimization of neural networks benefits from momentum \cite{sutskever2013importance}.
\vspace{-4mm}
\subsection{The Adaptive Gradient Family}
\vspace{-2mm}
The adaptive gradient methods have emerged as the most popular tool for training deep neural networks, and they have been of great use both industrially and academically. The adaptive gradient family divides an update step by a running root mean square (RMS) of the gradients \cite{Duchi:2011:ASM:1953048.2021068, Tieleman2012_rmsprop, journals/corr/KingmaB14_adam}, which speeds up training by effectively rescaling the update to the order of $\lambda O(1)$. To express the adaptive gradient family, we write the generalized update rule:
\begin{align}\label{eq: generalized update rule}
    \theta_{t+1} = \theta_t - \lambda_t G_t \nabla \ell(\theta_{t}) + \mu_t K_t (\theta_t - \theta_{t-1}),
\end{align}
where $G_t$ and $K_t$ are preconditioning matrices. Unlike Ref.~\cite{wilson2017marginal}, we do not assume $G_t$ and $K_t$ are dependent on each other. \rmsprop\ uses $\mu_t=0$ and
\begin{equation}
    G_t= \text{diag}\left((1-\nu)\sum_{i=0}^t \nu^{t-i} g_t \circ g_t\right)^{-\frac{1}{2}},\qquad g_t:=\nabla \ell(\theta_{t}),\qquad 0<\nu<1,
\end{equation}
so that the steps in SGD get divided by a RMS of the gradients, where $\nu$ is the adaptivity hyperparameter. ($\mu,\ \nu$ are also referred to as $\beta_1,\ \beta_2$ in literature.) The most widely used method, the \adam\ algorithm \cite{journals/corr/KingmaB14_adam}, incorporates momentum and also uses bias correction terms $c_m, c_n$ to correct the bias in the initialization. For completeness, we first write its update rule in its most familiar form:
\begin{align}\label{eq: adam}
    m_t= \mu m_{t-1} + (1-\mu) \nabla \ell(\theta_{t}), \quad
    n_t = \nu n_{t-1} + (1-\nu) (\nabla \ell(\theta_{t}))^2, \quad
    \theta_{t+1} = \theta_{t} - \lambda_t \frac{ m_t }{c_m \sqrt{n_t/c_n}},
\end{align}
where $(\nabla \ell(\theta_{t}))^2$ is computed element-wise, and $c_m:=\frac{1}{1-\mu^t}$ and $c_n:=\frac{1}{1-\nu^t}$, and $m_0=n_0=0$. Using the framework in Equation~\ref{eq: generalized update rule}, \adam\ can be expressed by
\begin{equation}
    H_t:= \text{diag}\left(\frac{1-\nu}{1 - \nu^t}\sum_{i=0}^t \nu^{t-i} g_t \circ g_t\right)^{\frac{1}{2}},\quad G_t = H_t^{-1},\quad K_t= H_t^{-1}H_{t-1},\qquad 0<\mu,\,\nu<1,
\end{equation}
where we have neglected factors $c_m$ and $(1-\mu)$ for clarity. The update rule thus becomes
\begin{align}\label{adam matrix expression}
    \theta_{t+1} = \theta_t - \lambda_t H_t^{-1} \nabla \ell(\theta_{t})
    + \mu_t \boxed{H_t^{-1} H_{t-1}}(\theta_t - \theta_{t-1}),
\end{align}
showing that the momentum term is reweighted by the factor in the rectangular box. The reason why \adam\ has $K_t= H_t^{-1}H_{t-1}$ is because \adam\ first computes an exponential average of past gradients and then divides the averaged gradient by the current preconditioner. Therefore, if we look at the parameter space, we see that the momentum of the parameter in the previous step, i.e.~$\theta_t - \theta_{t-1}$, is reweighted by the new preconditioner via $H_t^{-1} H_{t-1}$, and thus the parameter-space momentum and the adaptivity are coupled. This is drastically different from the situation of SGD, where an accumulated momentum in parameter space always has additive effects and is not rescaled on future steps.
\vspace{-4mm}
\subsection{\laprop}
\vspace{-2mm}
Intuitively, the term $H_t^{-1} H_{t-1}$ in Equation~\ref{adam matrix expression} may have negative effects in learning. The term $H_t^{-1} H_{t-1}$ reweights the momentum by the current gradient, which can be problematic if the current gradient is pathologically large, e.g.~due to noise, out-of-distribution data points or gradient explosion, and may cause the accumulated momentum to immediately vanish and hinder learning. As an alternative to \adam, we propose \laprop, which simply puts $K_t=I$, i.e.~the identity, and uses the following update rule:
\begin{align}
    \theta_{t+1} = \theta_t - \lambda_t H_t^{-1} \nabla \ell(\theta_{t}) 
    + \mu_t (\theta_t - \theta_{t-1}).
\end{align}
As shown, \laprop\ leaves the momentum of the parameter motion untouched, bringing it closer to the original momentum in non-adaptive methods. Then, even if a gradient is pathologically large, the effect of the gradient on the accumulated momentum is upper bounded by $\frac{1 -\mu}{\sqrt{1-\nu}}$ by construction (see Proposition~\ref{theo: laprop bound}), which suggests that \laprop\ puts more importance on its momentum than \adam, and that a single exploding gradient is not sufficient for \laprop\ to change its direction of optimization. Therefore, we conjecture that \laprop\ is better at overcoming barriers, and is more stable and harder to be trapped by sharp minima. One may doubt whether such a method really converges. In section \ref{sec: theory}, we rigorously show that \laprop\ indeed converges under the same assumption which \adam\ requires for convergence. The \laprop\ algorithm is 
\begin{wrapfigure}[11]{r}{0.44\textwidth}
\vspace{-8mm}
\begin{minipage}{0.44\textwidth}
\begin{algorithm}[H]
	\caption{\label{alg:simple laprop}\laprop\ }
	\begin{algorithmic}
		\STATE {\bfseries Input:} $\theta_1 \in \mathbb{R}^d$, learning rate $\{\lambda_t\}_{t=1}^T$,\linebreak decay parameters $0 \leq \mu,\nu < 1$, and $\epsilon\ll1$, bias correction factors $0<c_n, c_m<1$. Set $m_{0} =n_{0}= 0$.
		\FOR{$t=1$ {\bfseries to} $T$}
		%\STATE Draw a sample $s_t$ from $\mathbb{P}$.
		
		\STATE $g_{t} = \nabla_\theta \ell(\theta_{t})$
		\STATE $n_t = \nu n_{t-1} + (1-\nu) g_{t}^2$
		\STATE $m_t = \mu m_{t-1} + (1-\mu) \frac{ g_{t}}{\sqrt{n_t/c_n} + \epsilon} $
		%\STATE \#\#Following lines are the differences
		%\STATE $\mu_t = \beta_2 \mu_{t-1} + (1 - \beta_2) g_{t}$
		
		\STATE $\theta_{t+1} = \theta_t - \lambda_t m_t / c_m $
		\ENDFOR
	\end{algorithmic}
\end{algorithm}
\end{minipage}
\end{wrapfigure}
given as Algorithm \ref{alg:simple laprop}.

The idea of the preserved momentum in \laprop\ may be related to a couple of recent works which involve parameter-space operation on neural networks \cite{decouplingWeightDecay,averagingWeights,lookahead,lossSurface}. In those works, the parameters are directly averaged or decayed, and faster, more stable and more generalizable results are obtained. While the momentum mechanism in \adam\ is designed to average the gradient, \laprop\ averages the parameter updates throughout the training process, which is in agreement with the recent discovery of the effectiveness of parameter-space averaging. 

We recently noticed that Ref.~\cite{zou2018weighted} coincidentally proposed using the same momentum mechanism as used in \laprop; however, the authors did not examine the difference compared with \adam, and they considered a different preconditioner. \laprop\ can also be regarded as \rmsprop\ with momentum, which is not formally proposed or defined yet, but is already implemented by many deep learning libraries and used in a few works, and \laprop\ also uses the bias correction factors $c_m$ and $c_n$.
\vspace{-4mm}
\section{Interpolation Between Adaptive and Signed Gradient Methods}\label{sec: interpolation}
\vspace{-2mm}
An immediate consequence of \laprop\ is that we can tune $\nu\to0$ to recover the signed stochastic gradient (SSG) methods \cite{bernstein2018signsgd}, where the magnitude of the gradient is ignored and only the sign is used. Specifically, for $\nu=0$, we have $n_t=g_t^2$ and $m_t=\mu m_{t-1}+(1-\mu)\frac{g_t}{|g_t|}$ in Algorithm \ref{alg:simple laprop}, and the \laprop\ update rule becomes $\theta_{t+1} = \theta_t - (1-\mu)\lambda_t \textup{sgn}[\nabla \ell(\theta_{t})] + \mu (\theta_t - \theta_{t-1})$. However, the same is generally not true for \adam.
\vspace{-4mm}
\subsection{Divergence of Adam}
\vspace{-2mm}
If we set $\nu=0$ and $\mu\neq0$ for \adam, \adam\ will diverge. To see how it happens, we assume that the gradients $g_t$ are i.i.d.~drawn from a normal distribution, and for \adam\ we have
\begin{align}
    \text{Var}\left[\lim_{\nu\to 0}\frac{m_t}{\sqrt{n_t}}\right] = \text{Var}\left[\frac{\sum_i^t \mu^{t-i}g_i}{g_t}\right] \geq \text{Var}\left[\mu \frac{g_{t-1}}{g_t}\right] = \infty \qquad \text{  for }  t>1,
\end{align}
and thus one cannot divide the accumulated gradient by the current preconditioner as done by \adam. Nevertheless, \adam\ can still recover SSG for the special case of $\mu=\nu=0$. The behaviors of \adam\ and \laprop\ for the limiting values of $\mu$ and $\nu$ are summarized in Table~\ref{tab:interpolation}.
\begin{table}[t]
    \centering
    \vspace{-1mm}
    \caption{\label{tab:interpolation}The relationship of \adam\ and \laprop\ to other methods, where SSG-M represents SSG with momentum, and we refer to the bias-corrected version of \rmsprop. In all the limiting cases of $\mu$ and $\nu$, \laprop\ asymptotes to reasonable algorithms, while \adam\ does not for $\nu\to0$ and $\mu >0$. This table also shows that the signed gradient family are special cases of \laprop. We emphasize that although \adam\ and \laprop\ become asymptotically equivalent at the limit of $\nu\to1$, they still show different behaviours even at $\nu=0.999$.}
    \begin{tabular}{c|c|cc}
         & param. &$\nu\to 0$ & $\nu\to 1$ \\
        \hline
         \multirow{2}{*}{\adam} &$\mu \to 0$  & SSG & $\rmsprop$ \\
         &$\mu \to 1$ & \textcolor{red}{\textit{divergence}} & \adam$(\mu, \nu)$ \\
         \hline
        \multirow{2}{*}{\laprop} &$\mu \to 0$  & SSG & $\rmsprop$ \\
         &$\mu \to 1$ & SSG-M& \adam$(\mu, \nu)$
    \end{tabular}
    \vspace{-3mm}
\end{table}

The above suggests that $\mu$ and $\nu$ have complex relation in \adam. As $\mu$ controls the averaging of the past gradients on the numerator and $\nu$ controls the averaging on the denominator, if $\mu$ and $\nu$ do not match, \adam\ may diverge. This issue also appears in the convergence analysis of \adam\ \cite{Reddi2018convergence,da2018general}. For $\sqrt{\nu} > \mu$, we can prove that an update step of \adam\ is upper bounded by $\frac{1}{1-\mu/\sqrt{\nu}}$ (Proposition~\ref{theo: adam bound}), but for $\sqrt{\nu} < \mu$, we are not aware of a bound. In contrast, the update of \laprop\ is always upper bounded by $\frac{1}{\sqrt{1-\nu}}$ (Proposition~\ref{theo: laprop bound}). The same is demonstrated in Figure~\ref{fig:adam-divergence}, where \adam\ diverges on a simple task if $\nu$ and $\mu$ are not set appropriately. Similarly, we expect that \adam\ becomes unstable with a decreasing $\nu$, especially if the past gradients are noisy, or if the past gradients do not come from the same distribution as the learning proceeds.
\vspace{-4mm}
\subsection{Advantage of Interpolating to Signed Stochastic Gradient Methods}
\vspace{-2mm}
SSG uses the simple update rule $\theta_{t+1} = \theta_t - (1-\mu)\lambda_t \textup{sgn}[\nabla \ell(\theta_{t})] + \mu (\theta_t - \theta_{t-1})$.\footnote{Ref.~\cite{bernstein2018signsgd} proposed a slightly different version which averages the gradient as the momentum and then takes its sign. We experimentally find that our version is qualitatively the same as theirs.} This simple method is surprisingly efficient in training neural networks and it can be comparable to SGD and the adaptive methods, as shown in Ref.~\cite{bernstein2018signsgd}. Especially, it is found to be effective when the signal-to-noise ratio is low, which is also evidenced by our experiments in section~\ref{sec: rosenbrock}. Furthermore, Ref.~\cite{balles2017dissecting} suggests that the sign of the gradient actually accounts for most of the improvement offered by the adaptive gradient methods, and that the adaptive gradient methods may be interpreted as using the variance of the gradient to adaptively rescale the step in SSG. Therefore, it is reasonable to consider an interpolation from the adaptive methods to the SSG methods, i.e.~to gradually tune off the adaptivity. As we have shown, only \laprop\ can realize such an interpolation in general.

The interpolation to SSG offers two advantages for \laprop. The first advantage is that with the presence of the interpolation, \laprop\ always has stable and reasonable behaviour for all different hyperparameter settings. In our various experiments, we notice that if fine-tuning is needed, the fine-tuning of \laprop\ is easier than that of \adam, and we see that \laprop\ changes stably and smoothly regarding its varying hyperparameters, as in section~\ref{sec: rosenbrock} and \ref{sec: neural style transfer}. The easier fine-tuning may save a great amount of effort if a large-scale task is involved. The second advantage is that since SSG is effective when the noise is large, \laprop\ may be tuned to the SSG side according to the noise level of the task and achieve higher performance or faster speed. For some tasks where the training is unstable only at the beginning, it is possible to use a small $\nu$ at the beginning and change to $\nu\to1$ later to obtain faster final convergence.

In summary, \laprop\ decouples the parameter-space momentum from the adaptivity so that the effects of $\mu$ and $\nu$ are independent of each other, which leads to more flexibility, stability and better performance, and gives the optimizer more tunable behaviour than \adam. The default \laprop\ hyperparameters we recommend are $\lambda=4\times10^{-4}$, $\mu= 0.8$ - 0.9, $\nu=0.95$ - 0.999 and $\epsilon=10^{-15}$. Nevertheless, hyperparameter settings of \adam\ always work for \laprop. If the task is noisy or complex, one may try $\nu=0$~-~0.9 in the beginning and probably increase $\nu$ to a larger value later. \laprop\ is found to be highly stable and it works across different hyperparameter settings, and it is less sensitive to its hyperparameter settings when compared with \adam.
\vspace{-4mm}
\section{Mathematical Properties and Convergence Analysis}\label{sec: theory}
\vspace{-2mm}
The most important property of \laprop\ is probably that its update is always bounded: 
\begin{proposition}\label{theo: laprop bound}
Bound for \laprop\ update. Let $m_t$ be defined as in Algorithm~\ref{alg:simple laprop}, and set $c_n=1 - \nu^t$, $c_m = 1 - \mu^t$. Then the magnitude of the update is bounded from above as $| \frac{m_t}{c_m}| \leq \frac{1}{\sqrt{1-\nu}}$. 
\end{proposition}
An important feature of this bound is that it only depends on $\nu$. This is in contrast with the analysis for \adam:
\begin{proposition}\label{theo: adam bound}
Bound for Adam update. Let $m_t,\ n_t,\ c_n,\ c_m$ be defined as in Equation~\ref{eq: adam}, and set $c_n=1 - \nu^t$, $c_m = 1 - \mu^t$. Assume $\mu < \sqrt{\nu}$. Then the magnitude of the update is bounded from above as $| \frac{ m_t }{c_m \sqrt{n_t/c_n}}| \leq  \frac{1}{1 -\gamma}$, where $\gamma = \frac{\mu}{\sqrt{\nu}}$.
\end{proposition}
The proofs are given in the appendix. Notice that there are two key points: (1) the bound depends on the the ratio $\frac{\mu}{\sqrt{\nu}}$, suggesting that the momentum $\mu$ and the adaptivity $\nu$ are coupled; (2) the bound only applies when $\mu > \sqrt{\nu}$, suggesting that the range of $\mu$ and $\nu$ is limited. Although the default setting of \adam\ is $(\mu=0.9,\ \nu=0.999)$ which is within the bound, yet as our experiments have demonstrated, with such a restriction removed, \laprop\ can potentially achieve much more. 

Next, we present the convergence theorem of \laprop\ in a convex setting. The proof closely follows the results of the \adam\ style optimizers in Ref.~\cite{journals/corr/KingmaB14_adam, Reddi2018convergence}, and most of the derivations thereof can be readily applied to \laprop. Note that a rigorous proof of convergence for the adaptive gradient family has been a major challenge, with various bounds and rates present. In this section, we aim to present a bound whose terms are qualitatively insightful when compared with those of \adam.

\begin{theorem}\label{theo: regret bound}
(Regret bound for convex problem) Let the loss function be convex and the gradient be bounded, with $||\nabla \ell_t(\theta)||_\infty \leq G_\infty$ for all $\theta\in \mathbb{R}^d$, and let the distance between any $\theta_t$ learned by \laprop\ be bounded, with $||\theta_{t_1} - \theta_{t_2}||_\infty \leq D$, and let $\mu,\ \nu \in [0,1)$. Let the learning rate and the momentum decay as $\lambda_t=\lambda/\sqrt{t},\ \mu_t=\mu\zeta^{t-1}$ for $\zeta\in (0, 1)$. Define $s_t:=\sqrt{\frac{n_t}{c_n}}$. If we assume $\frac{s_{t+1}}{\lambda_{t+1}}\geq\frac{s_{t}}{\lambda_{t}}$, then the regret $R(T):=\sum_t^T (\ell(\theta_t)- \ell(\theta^*))$ can be bounded from above as
\begin{equation}
\begin{split}
     R(T) \leq &\frac{D^2 \sqrt{T}}{2\lambda(1-\mu)}\sum_{i=1}^d s_{T,i} + \frac{\mu d D^2 G_\infty}{2\lambda(1-\mu)(1-\zeta)^2} +\frac{\lambda\sqrt{1 + \ln{T}}}{(1-\mu)(1-\nu)\sqrt{1-\nu}}\sum_{i=1}^d ||g_{1:T,i}||_2,
\end{split}
\end{equation}
where in the worst case, we have $R(T) \leq \frac{D^2 \sqrt{T}}{2\lambda(1-\mu)}\sum_{i=1}^d s_{T,i} + \frac{\mu d D^2 G_\infty}{2\lambda(1-\mu)(1-\zeta)^2} +\frac{2\lambda d G_\infty\sqrt{T}}{(1-\mu)(1-\nu)}$.
\end{theorem}
We see that the major difference between this bound and that of \amsgrad\ in Ref.~\cite{Reddi2018convergence} lies in the third term, where \laprop\ replaces the factor $\frac{1}{(1-\mu)(1-\gamma)}$ by $\frac{1}{1- \nu}$.\footnote{Compared with Ref.~\cite{Reddi2018convergence}, the second term in the bound is refined and it does not include the square of $\frac{1}{1-\mu}$.} \laprop\ converges for any $\nu\in [0,1)$, while the variants of \adam\ depends on the relation between $\mu$ and $\nu$, i.e.~$\sqrt{\nu} > \mu$ must hold true. As a side note, the assumption $\frac{s_{t+1}}{\lambda_{t+1}}-\frac{s_{t}}{\lambda_{t}}\geq0$ is crucial, which also appears in the proof for the convergence of \adam\ and other known adaptive optimizers \cite{journals/corr/KingmaB14_adam,chen2018convergence}. This assumption is shown to be necessary for the convergence of \adam\ \cite{Reddi2018convergence}, and we find that \laprop\ also converges provided with such an assumption.

From the above theorem, one see that the average regret of \laprop\ converges at the same asymptotic rate as SGD and other adaptive optimizers such as \adam\ and \amsgrad, i.e.~$\frac{R(T)}{T} \sim O\left(\frac{1}{\sqrt{T}}\right)$.
Like \adam\ and \amsgrad, the above regret bound can be considerably better than $O(\sqrt{d T})$ if the gradient is sparse, as shown in Ref.~\cite{Duchi:2011:ASM:1953048.2021068}. The experimental fact that the adaptive gradient family is faster at training neural networks suggests that the gradient is often indeed sparse.
%\begin{table*}[t]
%    \centering
%    \caption{\label{tab:optimizer comparison}Classification of different %optimization algorithms according to Equation~\ref{eq: generalized update %rule}, with $I$ denoting the identity matrix. \laprop\ differs from \adam\ by %how the momentum is computed, and differs from the SGD-M by the way the %gradient is handled. This table also shows that the way \adam\ computes its %momentum is not quite in line with other methods.}
%    \begin{tabular}{cc|cc|c|cc}
%    & &\multicolumn{2}{c|}{\textup{non-adaptive}} & adaptive & %\multicolumn{2}{c}{adaptive \& momentum}\\
%         & & SGD & SGD-M & RMSProp & ADAM & \laprop\ \\
%         \hline
%    \multirow{2}{*}{gradient computation} &     $\lambda_t$ & $\lambda$ & %$\lambda$ & $\lambda$ &  $\lambda\frac{1-\mu}{1-\mu^t}$ & %$\lambda\frac{1-\mu}{1-\mu^t}$\\
%        & $G_t$ & $I$ & $I$ & $H_t^{-1}$ & $H_t^{-1}$&  $H_t^{-1}$\\
%         \hline
%         
%    \multirow{2}{*}{momentum computation}&     $\beta_t$& $0$ & $\mu$& $0$ & %$\mu$ & $\mu$\\
%        & $K_t$ & $-$ & $I$ & $-$ & $H_t^{-1}H_{t-1}$ & $I$ \\
%    \end{tabular}
%\end{table*}

\begin{figure*}[tb!]
\begin{subfigure}[b]{0.250\textwidth}
    \centering
    \includegraphics[trim=0 0 0 1, clip, width=1\textwidth]{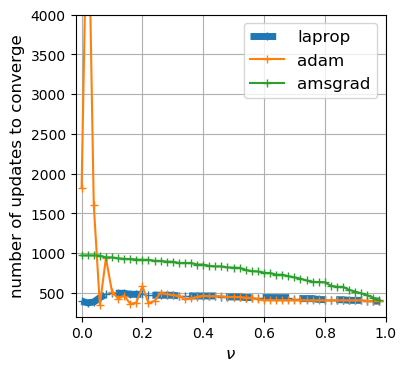}
    \vspace{-6mm}
    \caption{$\sigma=0.04$}
    \label{fig: rosen0.04}
\end{subfigure}
\hfill
\begin{subfigure}[b]{0.235\textwidth}
    \centering
    \includegraphics[trim=20 0 0 0, clip,width=1\textwidth]{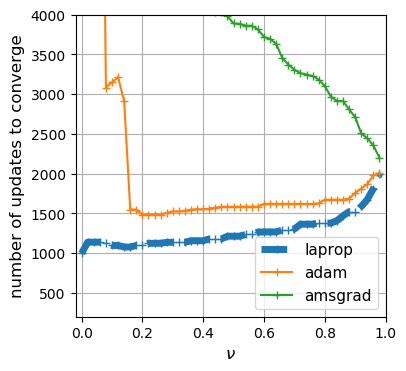}
    \vspace{-6mm}
    \caption{$\sigma=0.10$}
    \label{fig: rosen0.08}
\end{subfigure}
\begin{subfigure}[b]{0.240\textwidth}
    \centering
    \includegraphics[trim=20 0 0 0, clip, width=1\textwidth]{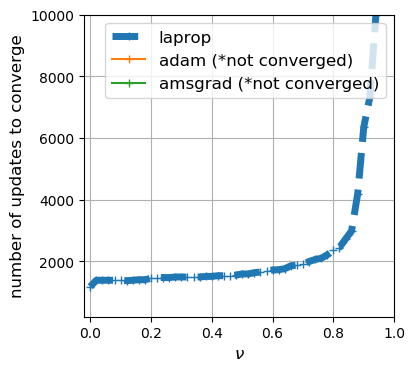}
    \vspace{-6mm}
    \caption{$\sigma=0.12$}
    \label{fig: rosen0.12}
\end{subfigure}
\hfill
\begin{subfigure}[b]{0.240\textwidth}
    \centering
    \includegraphics[trim=20 0 0 0, clip, width=1\textwidth]{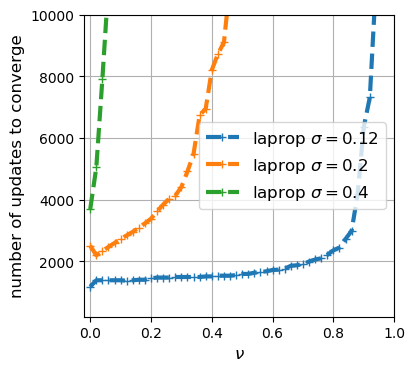}
    \vspace{-6mm}
    \caption{\laprop\ with $\sigma\geq 0.12$}
    \label{fig: rosen0.2}
\end{subfigure}
%\vspace{-3mm}
\caption{\label{fig: rosen}Time it takes to converge on the noisy Rosenbrock task plotted against $\nu$, with $\sigma$ being the noise level. (a) When the noise is small, the optimization speed of \laprop\ is almost invariant w.r.t.~$\nu$, demonstrating its flexibility in hyperparameters compared with the other optimizers; (b) when the noise gets larger, the performance of \adam\ and \amsgrad\ decreases, and they cannot work in the small $\nu$ regime where \laprop\ has its best performance; (c, d) For $\sigma \geq 0.12$, only \laprop\ converges even if we lengthen the optimization to $10000$ steps. Data points are plotted at equal intervals for all the curves, and we see \laprop\ is much stabler. Results for different learning rates for $\sigma=0.10$ are shown in the appendix.}
\vspace{-5mm}
\end{figure*}

\vspace{-4mm}
\section{Experiment}\label{sec: experiment}
\vspace{-1mm}
\subsection{Optimizing the Noisy Rosenbrock Loss}\label{sec: rosenbrock}
\vspace{-2mm}
In this section, we experimentally show the improved performance of \laprop\ compared with the previous methods, especially \adam. Detailed settings are given in the appendix. First, we consider optimization of the Rosenbrock loss function \cite{rosenbrock1960automatic}, for which the optimum is known exactly. Given two parameters $(x,y)$, the noisy Rosenbrock loss is defined as $\ell(x, y) = (1 -(x + \epsilon_1))^2 + 100\left((y + \epsilon_2)-(x + \epsilon_1)^2\right)^2$,
where $\epsilon_1$ and $\epsilon_2$ are noise terms. At each update step, $\epsilon_1$ and $\epsilon_2$ are independently sampled from $Uniform(-\sigma, \sigma)$, and thus the loss landscape is effectively shifted by $\epsilon_1$ and $\epsilon_2$. Parameters $x, y$ are initialized to be $0$, and the optimal solution is $(1, 1)$. When $(x,y)$ is sufficiently optimized such that $(1 -x)^2 + 100\left(y-x^2\right)^2<0.1$ holds, we say that the optimization is successful and it has converged. As shown in Figure~\ref{fig: rosen}, \adam, \amsgrad\ and \laprop\ have different behaviors on this simple task. When the noise is small, \laprop\ is one of the fastest method, but when the noise gets large, the task becomes increasingly difficult and only \laprop\ can work. It is worth noting that \laprop\ is more stable and it is less sensitive to the hyperparameter $\nu$. Its performance changes with $\nu$ smoothly. Hyperparameters of the three optimizers are always identical in the experiment.
\vspace{-4mm}
\subsection{Neural Style Transfer}\label{sec: neural style transfer}
\begin{wrapfigure}[13]{r}{0.49\linewidth}%[t]
\vspace{-17mm}
\centering
\begin{subfigure}[b]{0.43\linewidth}
    \centering
    \includegraphics[trim=0 0 0 0, clip,width=1\textwidth]{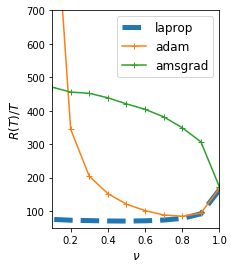}
    \vspace{-7mm}
    \caption{regret vs.~$\nu$}
\end{subfigure}
\hfill
\centering
\begin{subfigure}[b]{0.545\linewidth}
    \centering
    \includegraphics[trim=0 0 0 30, clip,width=1\textwidth]{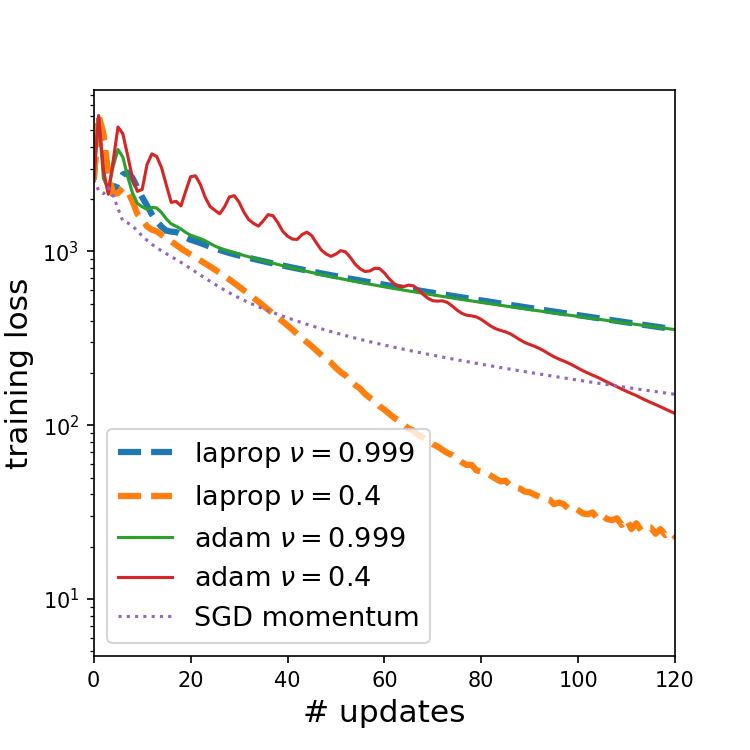}
    \vspace{-7mm}
    \caption{optimization trajectories}
\end{subfigure}
\vspace{-6mm}
\caption{\label{fig:neural style}Neural style transfer with different optimizers. (a) The average regret $R(T)/T$ at $T=1000$ plotted against $\nu$. A lower value corresponds to a better convergence rate \cite{journals/corr/KingmaB14_adam, Reddi2018convergence}. (b) Example optimization curves of different optimizers for the first 120 updates.}
\end{wrapfigure}
\vspace{-1mm}
Following the optimization task above, we consider the task of generating an image with a given content and a given style, which is also purely optimizational and it is described in detail in Ref.~\cite{gatys2015neural}. As shown in Figure~\ref{fig:neural style}(a), we still see \laprop\ performs best across different $\nu$ values. Although the trend in Figure~\ref{fig:neural style}(a) agrees well with that in section~\ref{sec: rosenbrock}, actually, the best $\nu$ value for $\laprop$ is non-zero and is found to be $0.4$, implying that the tuning of $\nu$ is nontrivial and may improve the performance. On this task, a smaller $\nu$ offers a better performance for \laprop\ but a worse performance for \amsgrad, and \adam\ performs well only when $\nu$ is relatively large. Examples of the optimization curves are shown in Figure~\ref{fig:neural style}(b). We see that while \laprop\ is stable and achieves its fastest speed for $\nu=0.4$, \adam\ exhibits oscillatory behavior. Results for different learning rates are provided in the appendix.

\vspace{-3mm}
\subsection{Training Extremely Deep Fully Connected Networks}
\begin{wrapfigure}[12]{r}{0.52\linewidth}%[t]
\vspace{-5mm}
\centering
\begin{subfigure}[t]{0.510\linewidth}
    \centering
    \includegraphics[trim=0 0 0 0, clip,width=1\linewidth]{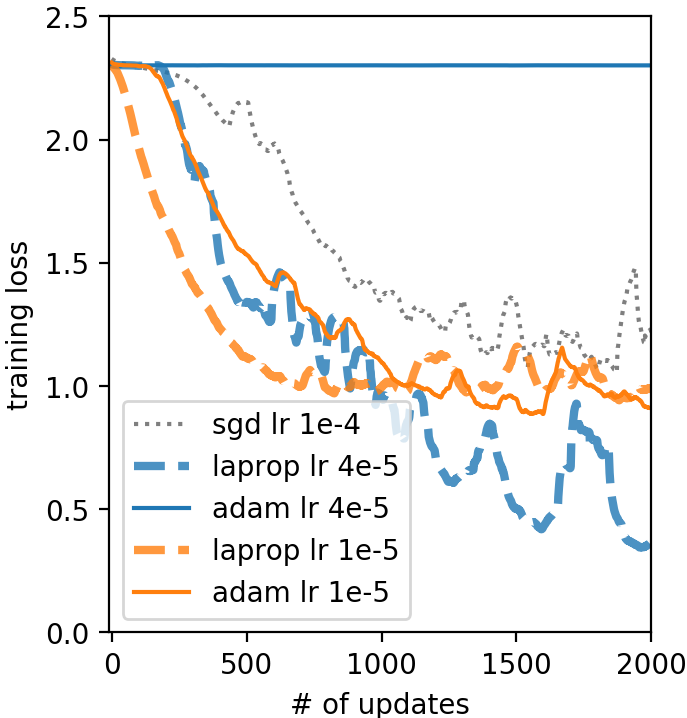}
    %\vspace{-7mm}
    \caption{depth 500}
    %\label{fig: depth, width256}
    %\vspace{-3mm}
\end{subfigure}
\begin{subfigure}[t]{0.475\linewidth}
    \centering
    \includegraphics[trim=35 0 0 0, clip,width=1\textwidth]{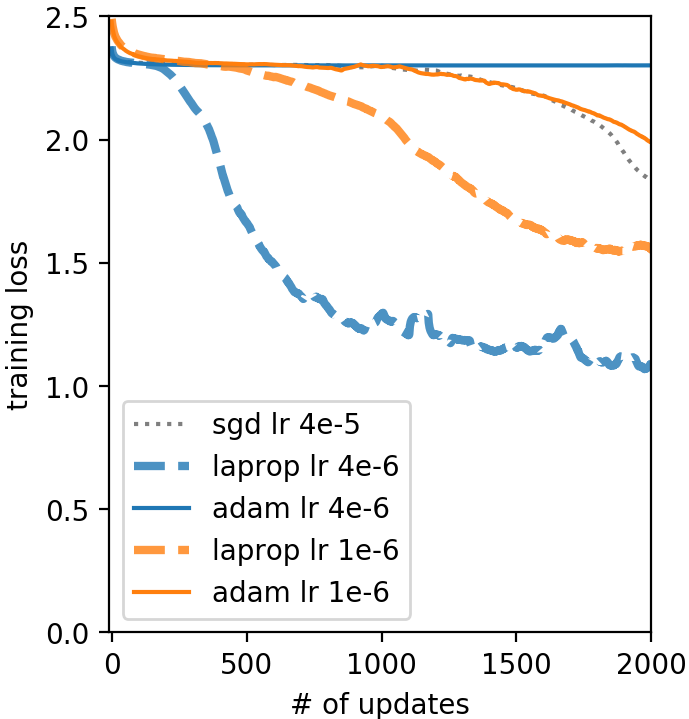}
    %\vspace{-7mm}
    \caption{depth 1000}
    %\label{fig: depth, width256}
    %\vspace{-3mm}
\end{subfigure}
\vspace{-2mm}
\caption{\label{fig: depth}Training curves of deep FC networks. }
\vspace{-2mm}
\end{wrapfigure}
\vspace{-1mm}
In the following, we demonstrate that \laprop\ can be used as a reliable substitute for \adam\ in deep learning, and it perform comparably to or outperforms \adam\ on a variety of tasks even for $\nu\sim1$, which is the default setting of \adam. First, we show that \laprop\ has stronger ability to deal with hard optimizations in deep learning. Specifically, \laprop\ can train extremely deep fully connected (FC) networks better than \adam\ does, which we demonstrate on MNIST using the ReLU activation. Ref.~\cite{hanin2018start} shows that it is hard to initialize the training of FC networks at extreme depths due to the gradient explosion or vanishing problem. Our results are shown in Fig.~\ref{fig: depth}. We use the default parameter settings of \laprop\ except for the learning rate, and the network width is set to be 256. We see that \laprop\ can learn with a larger learning rate compared with \adam, and \laprop\ reduces the training loss faster. The results imply that \laprop\ is more robust for handling complex and pathological problems than \adam.

\subsection{Translation and Pretraining with Transformers}
\begin{figure}[tb]
\begin{subfigure}[b]{0.49\textwidth}
    \centering
    \includegraphics[width=0.92\textwidth]{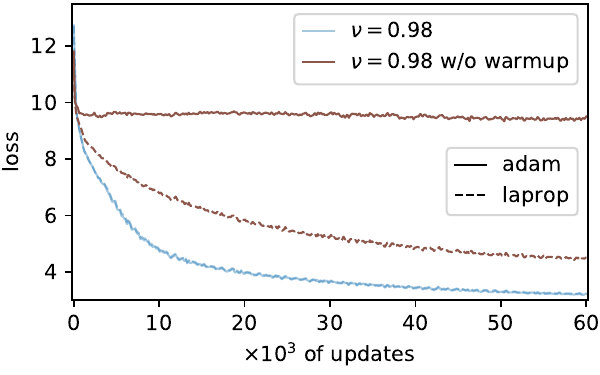}
    \vspace{-2mm}
    \caption{training of IWSLT14 de-en}\label{fig: IWSLT}
\end{subfigure}\hfill
\begin{subfigure}[b]{0.49\textwidth}
    \centering
    \vspace{2mm}
    \includegraphics[width=0.92\textwidth]{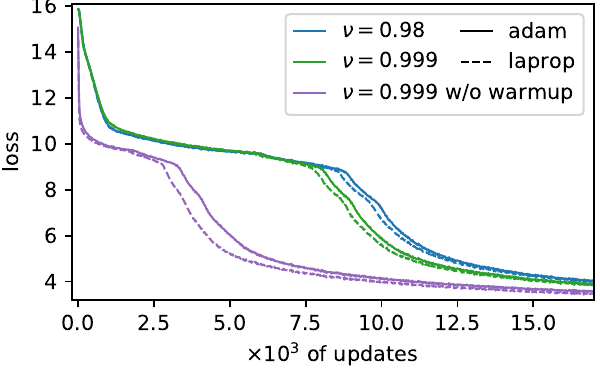}
 \vspace{-2mm}
    \caption{training RoBERTa on the full English Wikipedia}\label{fig: Roberta}
\end{subfigure}
\vspace{-2mm}
\caption{Learning curves of the transformer tasks. When there is a warmup, the learning rate linearly increases from zero to the maximum and then decreases; otherwise it starts from the maximum and decreases. The warmup includes the first $2\times 10^3$ updates in (a), and $10\times 10^3$ updates in (b).}\label{fig: transformer}
\vspace{-3mm}
\end{figure}

\vspace{-1mm}
As a recently proposed modern network architecture \cite{vaswani2017attention}, the transformer is known to be difficult to train. Especially, it usually requires a warm-up schedule of learning rates if a large learning rate is to be used, and it is typically trained by \adam\ with $\nu\leq0.999$. We show that the performance and the optimization speed of \laprop\ parallel or outperform those of \adam. We consider the tasks of IWSLT14 German to English translation \cite{IWSLT14} and the large-scale RoBERTa(base) pretraining using the full English Wikipedia \cite{liu2019roberta}, which are of both industrial and academic interest. We follow the prescriptions given by Ref.~\cite{ott2019fairseq} and use the default network architectures and settings thereof, and the results are shown in Figure \ref{fig: transformer}. For the IWSLT14 translation task, when a warmup is used, the learning curves of \adam\ and \laprop\ coincide; without a warmup, as is also shown in Ref.~\cite{liu2019variance}, \adam\ gets trapped by a bad minimum and does not converge to a low loss, while we find that \laprop\ does not get completely trapped and it continues looking for a better solution, indicating its stronger capability of overcoming barriers. For the RoBERTa pretraining, due to limited computational resources, we do not carry out the whole pretraining process and we only do the initial $2\times10^4$ updates. We find that in our setting, we can use $\nu=0.999$ as well as the default $\nu=0.98$, and we can safely ignore the warmup, and in all the cases, the speed of \laprop\ is faster than \adam, and when the optimization speed of the problem is faster, \laprop\ outperforms \adam\ even more. We also notice that a smaller $\nu$ speeds up the training at the initial stage while a larger $\nu$ speeds up later, which implies that a $\nu$ schedule may be beneficial. Details are given in the appendix.

\vspace{-2mm}
\subsection{Reinforcement Learning of Atari Games}
\vspace{-1mm}
\begin{figure}[bt]
\centering
\begin{subfigure}[b]{0.42\textwidth}
    \centering
    \includegraphics[width=\textwidth]{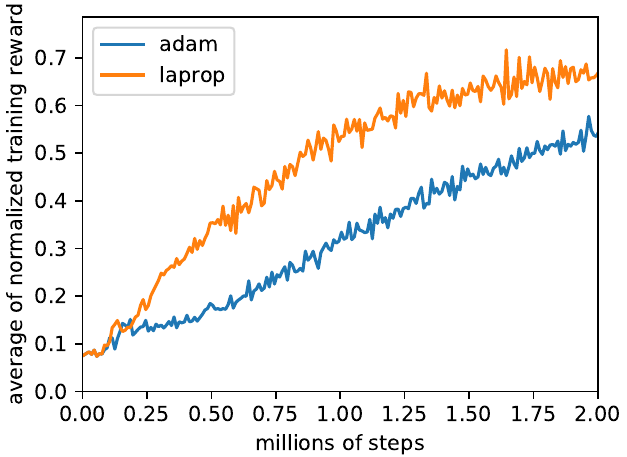}
    \vspace{-7mm}
    \caption{\label{fig: atari total}}
\end{subfigure}
\qquad
\begin{subfigure}[b]{0.42\textwidth}
    \centering
    \includegraphics[width=\textwidth]{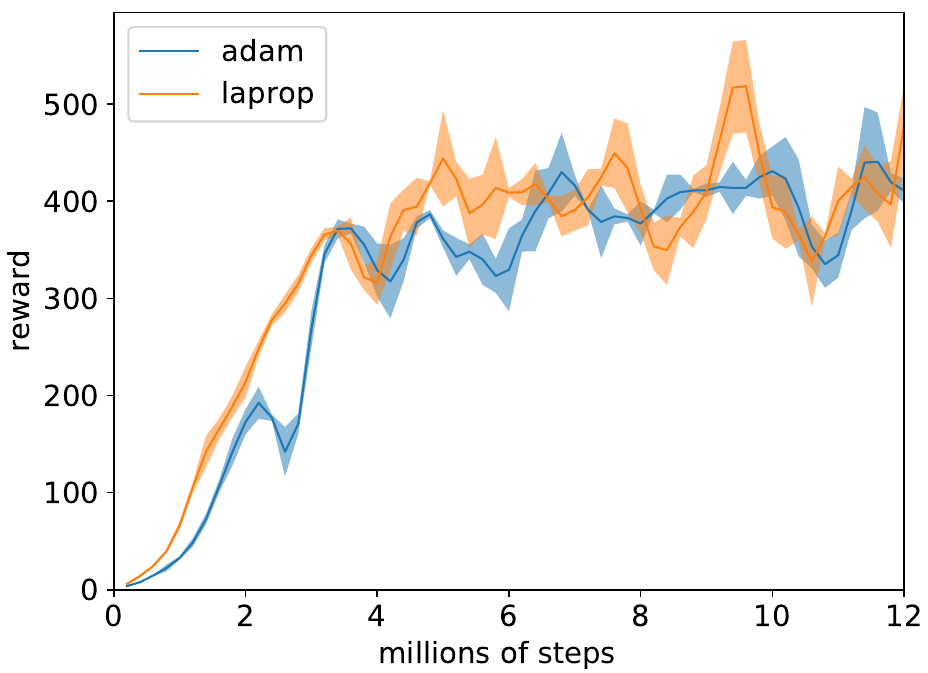}
    \vspace{-7mm}
    \caption{\label{fig: breakout}}
\end{subfigure}
\caption{\label{fig: RL}(a) Average of the normalized training performances on 20 different Atari2600 games, for which the learning shows clear progress at early stages. (b) Test performance on Atari2600 game {\it Breakout}. For every 0.2 million steps, the trained model is evaluated for 20 times to obtain a test performance, which is plotted with the standard error. Gaussian smoothing with a standard deviation of 1 is applied to the nearby data points of the evaluated avrage test performance.}
\vspace{-5mm}
\end{figure}
We compare the performance of \laprop\ and \adam\ on a modern reinforcement learning (RL) task, i.e.~learning to play Atari2600 games, and we use the Rainbow DQN algorithm in Ref.~\cite{rainbow}. As reported in the original paper Ref.~\cite{rainbow}, the full training of all of the games is extremely time-consuming, which shows the difficulty of the task. Here, a few minor settings of the default Rainbow DQN have been modified to accelerate training. Since the task is difficult, we investigate the final performance of training only on a single game, and we compare the early-stage performance of \laprop\ and \adam\ on other games. The results are shown in Figure~\ref{fig: RL}. The training performances of \laprop\ and \adam\ are averaged on 20 different Atari2600 games after being normalized, shown in Figure~\ref{fig: RL}(a). The evaluated test performance on the Atari2600 game {\it Breakout} is shown in Figure~\ref{fig: RL}(b). Details and the results of all the 20 games are given in the appendix. We find that \laprop\ always starts learning earlier than \adam\ and often achieves high performances faster than \adam. Its performance improves with fluctuation just like \adam, but its overall performance is better. It is important to realize that the fluctuation of performance is a consequence of RL and is not merely due to noise. For the game {\it Breakout}, the best trained model obtained by \laprop\ in training has a test performance of $607.3\pm47.6$, which is higher than that of \adam, which is $522.9\pm40.6$. Therefore, we see that \laprop\ can be used to achieve both faster and better RL results. Concerning RL behavior, in our experiments, we also notice that \laprop\ is qualitatively more aggressive than \adam\ in learning and exploration.

\vspace{-4mm}
\section{Conclusion}
\vspace{-2mm}
Motivated by a series of previous works, we have proposed an optimization method that is shown to be effective for training modern neural networks on various tasks. While the proposed method does outperform and show better flexibility than other members in the adaptive gradient family, we remark that the understanding of its true advantage needs to be tested with more experiments and be investigated in further detail. Additional experimental results on CIFAR10 using residual networks and on IMDB using LSTM are provided in the appendix.
\clearpage

\bibliography{example_paper}
\bibliographystyle{unsrt}

\clearpage
\appendix 
\onecolumn
%\section{Broader Impact Statement}
%This work aims at improving a fundamental algorithm in deep learning; therefore, its direct impact is the acceleration of training of various neural network based model. Such acceleration is beneficial in general, since training deep neural networks on hard tasks can take very long time and consumes abundant resources and electricity in particular. Therefore, we believe advancement in fundamental algorithms may also have some impact to society.

\section{Mathematical Proofs}
\subsection{Proof of Proposition~1}
\vspace{-2mm}
Expand the update step:
\begin{equation}
    \begin{split}
    |m_t| &\leq \mu |m_{t-1}| + (1-\mu) \frac{|g_t|}{\sqrt{n_t/c_n(t)}}\\
    &\leq \mu |m_{t-1}| + (1-\mu) \frac{|g_t|}{\sqrt{[(1-\nu)g_t^2 + \nu n_{t-1}]/c_n(t)}}\\
    &\leq \mu |m_{t-1}| + (1-\mu)\frac{|g_t|}{\sqrt{[(1-\nu)g_t^2]/c_n(t)}}\\
    &\leq \mu |m_{t-1}| + \frac{1-\mu}{\sqrt{1 - \nu}}
    \end{split}
\end{equation}
and this defines a recurrence relation that solves to
\begin{align}
    |m_t| \leq \frac{1 - \mu^t}{\sqrt{1 - \nu}}
\end{align}
whereby
\begin{align}
    \frac{|m_t|}{c_m(t)} \leq  \frac{1}{\sqrt{1 - \nu}}.
\end{align}

\subsection{Proof of Proposition~2}
\vspace{-2mm}
Expand the term:
\begin{equation}
    \begin{split}
    \frac{|m_t|}{\sqrt{n_t}} &\leq \mu \frac{|m_{t-1}|}{\sqrt{n_t}} + (1-\mu) \frac{|g_t|}{\sqrt{n_t}}\\
    & \leq \mu \frac{|m_{t-1}|}{\sqrt{\nu n_{t-1}}} + (1-\mu) \frac{|g_t|}{\sqrt{(1-\nu)g_t^2}}\\
    & \leq \frac{1-\mu}{\sqrt{1 -\nu}}\sum_{t'=0}^{t} \frac{\mu^{t'}}{\nu^{\frac{t'}{2}}}
    \end{split}
\end{equation}
and if $\mu < \sqrt{\nu}$, then
\begin{align}
    \frac{|m_t|}{\sqrt{n_t}} \leq \frac{1-\mu}{\sqrt{1 -\nu}} \frac{1}{1 - \gamma}
\end{align}
where $\gamma = \frac{\mu}{\sqrt{\nu}}$. Putting in the bias corrections, we obtain that
\begin{equation}
    \frac{|m_t|/c_m(t)}{\sqrt{n_t/c_n(t)}} \leq \frac{1}{\sqrt{1-\nu}}\frac{1}{1-\gamma},
\end{equation}
and we are done.

\subsection{Proof of Theorem~1}
\vspace{-2mm}
By convexity, we have
\begin{equation}
    \ell(\theta_t) - \ell(\theta^*) \leq \sum_{i=1}^d g_{t,i}(\theta_{t,i} - \theta_i^*),
\end{equation}
where $\theta^*$ is the global minimum of $\ell$. In the following, we focus on a single component with index $i$ and bound the term $g_{t,i}(\theta_{t,i} - \theta_i^*)$. We plug in the \laprop\ update rule and obtain
\begin{equation}
    \begin{split}
    \theta_{t+1} &= \theta_t - \lambda_t\frac{m_t}{c_m}\\
    &= \theta_t - \frac{\lambda_t}{c_m}\bigg( \mu_t m_{t-1} + (1 - \mu_t) \frac{g_t}{s_t}\bigg)
    \end{split}
\end{equation}
where we have defined $s_t := \sqrt{n_t / c_n}$. We subtract $\theta^*$ from both sides and square them to obtain
\begin{align}
    (\theta_{t+1} - \theta^*)^2 = (\theta_t - \theta^*)^2 - 2 \frac{\lambda_t}{c_m}\bigg(\mu_t m_{t-1} + (1 - \mu_t) \frac{g_t}{s_t}\bigg)(\theta_t - 
    \theta^*) + \frac{\lambda_t^2}{c_m^2}m_t^2.
\end{align}
The terms are rearranged to give obtain an expression for $g_t(\theta_t - \theta^*)$ using the second term above. In the second line below, we use the inequality $ab<\frac{a^2 + b^2}{2}$.
\begin{equation}
    \begin{split}
    g_{t}(\theta_t - \theta^*) &= \frac{c_m}{2\lambda_t(1-\mu_t)}\bigg[(\theta_t - \theta^*)^2 - (\theta_{t+1} -\theta^*)^2 \bigg]s_t + \frac{\mu_t}{1 - \mu_t} {s_t} m_{t-1}(\theta^* - \theta_t)+ \frac{\lambda_t}{2c_m(1-\mu_t)}s_t m_t^2\\
    &\leq \frac{c_m}{2\lambda_t(1-\mu_t)}\bigg[(\theta_t - \theta^*)^2 - (\theta_{t+1} -\theta^*)^2 \bigg]s_t  +  \frac{\lambda_t}{2c_m(1-\mu_t)}s_t m_t^2 \\
    & \quad\quad\quad\quad\quad\quad\quad\quad\quad + \frac{\mu_t}{2\lambda_{t}(1-\mu_t)}(\theta^* - \theta_t)^2 s_t + \frac{\lambda_{t}\mu_t}{2(1-\mu_t)}s_t m_{t-1}^2
    \end{split}
\end{equation}
Proposition~1 is applied to bound $m_t$.
\begin{equation}
    \begin{split}
     g_{t}(\theta_t - \theta^*) &\leq \frac{c_m}{2\lambda_t(1-\mu_t)}\bigg[(\theta_t - \theta^*)^2 - (\theta_{t+1} -\theta^*)^2 \bigg]s_t  +  \frac{\lambda_t}{2(1-\mu_t)(1-\nu)}s_t\\
     &\quad\quad\quad\quad\quad\quad\quad\quad + \frac{\mu_t}{2\lambda_{t}(1-\mu_t)}(\theta^* - \theta_t)^2 s_t + \frac{\lambda_{t}\mu_t}{2(1-\mu_t)(1-\nu)}s_t\\
     &\leq \frac{c_m}{2\lambda_t(1-\mu_t)}\bigg[(\theta_t - \theta^*)^2 - (\theta_{t+1} -\theta^*)^2 \bigg]s_t + \frac{\mu_t}{2\lambda_{t}(1-\mu)}(\theta^* - \theta_t)^2 s_t + \frac{\lambda_t}{(1-\mu)(1-\nu)}s_t
    \end{split}
\end{equation}
Now we are ready to bound $R(T)$, to do this, we sum over the index $i$ and the times steps from $1$ to $T$ (recall that each of $m,\ \theta,\ s$ carries an index $i$):
\begin{equation}
    \begin{split}
    R(T) &\leq \sum_{i=1}^d \sum_{t=1}^T \bigg\{ \frac{c_{m,t}}{2\lambda_t(1-\mu_t)}\bigg[(\theta_t - \theta^*)^2 - (\theta_{t+1} -\theta^*)^2 \bigg]s_t + \frac{\mu_t}{2\lambda_{t}(1-\mu)}(\theta^* - \theta_t)^2 s_t + \frac{\lambda_t}{(1-\mu)(1-\nu)}s_t \bigg\} \\
    &= \sum_{i=1}^d \frac{c_{m,1}}{2\lambda_1(1-\mu_1)}(\theta_{1,i} - \theta^*_i)^2 s_{1,i}
    + \sum_{i=1}^d \sum_{t=2}^T \frac{1}{2}(\theta_{t,i} - \theta^*_i)^2 \underbrace{\bigg( \frac{c_{m,t}s_{t,i}}{(1-\mu_t)\lambda_t} - \frac{c_{m,{t-1}}s_{t-1,i}}{(1-\mu_{t-1})\lambda_{t-1}} \bigg)}_{A}
     \\
    &\quad + \sum_{i=1}^d \sum_{t=1}^T \frac{\mu_t s_{t,i}}{2\lambda_t(1-\mu)}(\theta_i^* - \theta_{t,i})^2 + \sum_{i=1}^d \sum_{t=1}^T \frac{\lambda_t}{(1-\mu)(1-\nu)}s_{t,i}.
    \end{split}
\end{equation}
We focus on the term $A$. Using $\mu_t<\mu_{t-1}$, we have
\begin{equation}
    A\leq \frac{c_{m,t}s_{t,i}}{(1-\mu_{t-1})\lambda_t}-\frac{c_{m,{t-1}}s_{t,i}}{(1-\mu_{t-1})\lambda_t}=\frac{1}{1-\mu_{t-1}}\left(\frac{c_{m,t}s_{t,i}}{\lambda_t} - \frac{c_{m,{t-1}}s_{t-1,i}}{\lambda_{t-1}}\right).
\end{equation}
Also with $c_{m,{t-1}}\leq c_{m,t}\leq1$ and the assumption $\frac{s_{t-1,i}}{\lambda_{t-1}}\leq\frac{s_{t,i}}{\lambda_t}$, we have
\begin{equation}
    \begin{split}
\frac{c_{m,t}s_{t,i}}{\lambda_t} - \frac{c_{m,{t-1}}s_{t-1,i}}{\lambda_{t-1}}\geq0.
    \end{split}
\end{equation}
Therefore we obtain
\begin{equation}
    \begin{split}
R(T) &\leq \sum_{i=1}^d \frac{c_{m,1}}{2\lambda_1(1-\mu_1)}(\theta_{1,i} - \theta^*_i)^2 s_{1,i}
    + \sum_{i=1}^d \sum_{t=2}^T \frac{1}{2}(\theta_{t,i} - \theta^*_i)^2 \bigg( \frac{c_{m,t}s_{t,i}}{(1-\mu_{t-1})\lambda_t} - \frac{c_{m,{t-1}}s_{t-1,i}}{(1-\mu_{t-1})\lambda_{t-1}} \bigg)\\
    &\quad + \sum_{i=1}^d \sum_{t=1}^T \frac{\mu_t s_{t,i}}{2\lambda_t(1-\mu)}(\theta_i^* - \theta_{t,i})^2 + \sum_{i=1}^d \sum_{t=1}^T \frac{\lambda_t}{(1-\mu)(1-\nu)}s_{t,i}\\
    &\leq \sum_{i=1}^d \frac{c_{m,1}s_{1,i}}{2\lambda_1(1-\mu)}D^2
    + \sum_{i=1}^d \sum_{t=2}^T \frac{D^2}{2(1-\mu)}\bigg( \frac{c_{m,t}s_{t,i}}{\lambda_t} - \frac{c_{m,{t-1}}s_{t-1,i}}{\lambda_{t-1}} \bigg)+ \sum_{i=1}^d \sum_{t=1}^T \frac{\mu_t s_{t,i}}{2\lambda_t(1-\mu)}D^2
     \\
    &\quad  + \sum_{i=1}^d \sum_{t=1}^T \frac{\lambda_t}{(1-\mu)(1-\nu)}s_{t,i}\\
    &\leq \frac{D^2\sqrt{T}}{2\lambda(1-\mu)}\sum_{i=1}^d s_{T,i} + 
    \frac{\mu d D^2 G_\infty}{2\lambda(1-\mu)(1-\zeta)^2} + 
    \frac{\lambda}{(1-\mu)(1-\nu)} \underbrace{\sum_{i=1}^d \sum_{t=1}^T \frac{s_{t,i}}{\sqrt{t}}}_{B}.
    \end{split}
\end{equation}
Note that we have used $||\theta_{t_1} - \theta_{t_2}||_\infty \leq D$ and $s_{t,i} \leq G_\infty$, and the second last term is derived using $\frac{\mu_t}{\lambda_t}=\frac{\mu\zeta^t\sqrt{t}}{\lambda}\leq\frac{\mu t\zeta^t}{\lambda}$. We now try to bound the term $B$ above. The worst case bound can be found by using $s_{t,i} \leq G_\infty$ and $\sum_t^T 1/\sqrt{t} < 2\sqrt{T} - 1$. Thus the worst case bound is $B \leq 2d G_\infty\sqrt{T}$, which is the same as other adaptive gradient methods.

Another bound, which is tighter when the gradient is sparse, can be obtained using the Cauchy-Schwarz inequality as follows:
\begin{equation}
    \begin{split}
    \sum_{i=1}^d \sum_{t=1}^T \frac{s_{t,i}}{\sqrt{t}} &\leq \sum_{i=1}^d \sqrt{\sum_t \frac{n_{t,i}}{c_n}} \sqrt{\sum_t \frac{1}{t}} \\
    &\leq \sqrt{1 +\ln{T}} \sum_{i=1}^d \sqrt{\sum_t \frac{n_{t,i}}{c_n}} \\
    &=  \sqrt{1 +\ln{T}} \sum_{i=1}^d \sqrt{\sum_t^T \frac{\sum_j^t (1-\nu)\nu^{t-j}g_{j,i}^2}{c_n}}\\
    %&\leq  \sqrt{1 +\log T} \sum_{i=1}^d \sqrt{\sum_t^T \frac{\sum_j^T (1-\nu)\nu^{t-j}g_{j,i}^2}{c_n}} \\
    &=  \sqrt{1 +\ln{T}} \sum_{i=1}^d \sqrt{\sum_t^T \sum_j^t \nu^{t-j}g_{j,i}^2}\\
    &\leq \sqrt{\frac{1 +\ln{T}}{1 - \nu}} \sum_{i=1}^d ||g_{1:T, i}||
    \end{split}
\end{equation}
and so we obtain the desired bound
\begin{align}
    R(T) \leq \frac{D^2\sqrt{T}}{2\lambda(1-\mu)}\sum_{i=0}^d s_{T,i} + 
    \frac{\mu d D^2 G_\infty}{2\lambda(1-\mu)(1-\zeta)^2} + 
    \frac{ \lambda \sqrt{1 +\ln{T}}}{(1-\mu)(1-\nu)\sqrt{1 - \nu}}  \sum_{i=1}^d ||g_{1:T, i}||.
\end{align}
The assumption $\frac{s_{t+1,i}}{\lambda_{t+1}}\geq\frac{s_{t,i}}{\lambda_t}$ is crucial as discussed in Ref.~\cite{Reddi2018convergence}. In two ways we can make sure that this assumption is satisfied: one is to increase $\nu$ gradually towards $1$ during training, as proven in Ref.~\cite{Reddi2018convergence}; another is to define an \amsgrad\ version of the adaptive method by substituting $s_{t,i} = \max(\sqrt{\frac{n_{t, i}}{c_n}}, \sqrt{\frac{n_{t-1, i}}{c_n}})$. However, the \amsgrad\ ``fix" is found to be harmful to the performance, and we discourage using an \amsgrad\ version of \laprop\ unless the problem demands so, especially at an early stage of training. Empirically, one may also enlarge the batch size to suppress the fluctuation of $s_{t,i}$ so that better convergence can be obtained.

\section{Practical Concerns and Various Extensions}
\subsection{Tuning $\epsilon$}
\vspace{-2mm}
The tuning of $\epsilon$ is found to be important for stability and convergence of \adam\ in some difficult tasks, such as in Rainbow DQN and Transformer training \cite{rainbow,bert}, and a small $\epsilon$ may result in an unstable learning trajectory. However, we find that this is usually not the case for $\laprop$, and $\laprop$ almost always works well with a small $\epsilon$. The $\epsilon$ can be freely set to be $10^{-8}$, $10^{-15}$ or $10^{-20}$, as long as it is within the machine precision. On the other hand, if we consider $\epsilon$ as a term that helps the optimizer to converge in the presence of the assumption discussed in the last section, it can be tuned to a non-negligible value such as $10^{-4}$, which actually, slows down the optimization significantly. Therefore, we encourage to use a larger batch size, a larger $\nu$ or an \amsgrad\ version of \laprop\ to ensure final convergence, rather than to use a large $\epsilon$. 

As \laprop\ is considerably stable, we find that gradient clipping is unnecessary, and we do not use gradient clipping in our experiments.
\subsection{Tuning $\nu$}
\vspace{-2mm}
The tuning of $\nu$ is nontrivial and it is basically determined by trial and error. Generally, a smaller $\nu$ makes \laprop\ closer to the signed gradient methods and the maximal update made by \laprop\ becomes smaller, which may be beneficial in noisy settings, and a larger $\nu$ makes $n_t$ change more slowly, which may benefit fine-tuning of the model and the final convergence. From a statistical viewpoint, $\nu$ is used to estimate the second moment of the gradient on sampled data, and therefore if the model changes quickly with optimization updates, a large $\nu$ will result in a large bias in the second-moment estimation. In our experiments, we find that a smaller $\nu$ sometimes indeed improves the loss faster at the initial stage where the model changes quickly, while a larger $\nu$ improves faster later, such as in Fig.~\ref{fig: RobertaBeta2}. For MNIST and CIFAR10, a small $\nu$ trains slightly faster only for the initial hundreds of updates and this occurs only for a limited range of $\nu$, which may be due to the simplicity of the tasks. We notice that when the training is stable, a larger $\nu$ almost always makes the training faster, and therefore we believe that increasing $\nu$ to a larger value at a later stage is beneficial. The tuning of $\nu$ for large-scale and difficult tasks can potentially bring improved results and we leave it for future research.
\subsection{Weight Decay for \laprop: \laprop W}
\vspace{-2mm}
As suggested in Ref.~\cite{loshchilov2017fixing}, we implement the weight decay separately, which may also be called the \laprop W algorithm. The algorithm is given by Algorithm~\ref{alg:lapropW}. Note that we combine the learning rate and the momentum so that even in the presence of a changing learning rate, the momentum still represents the average of the different update steps.
\begin{algorithm}[]
	\caption{\laprop W}
	\label{alg:lapropW}
	\begin{algorithmic}
		\STATE {\bfseries Input:} $\theta_1 \in \mathbb{R}^d$, learning rate $\{\lambda_t\}_{t=1}^T$, weight decay $\{w_t\}_{t=1}^T$, decay parameters $0 \leq \mu < 1,\ 0 \leq \nu < 1 ,\ \epsilon \ll 1$, bias correction factors $0<c_n, c_m<1$. Set $m_{0} = 0$, $n_{0} = 0$.
		\FOR{$t=1$ {\bfseries to} $T$}
		%\STATE Draw a sample $s_t$ from $\mathbb{P}$.
		
		\STATE $g_{t} = \nabla_\theta \ell(\theta_{t})$
		\STATE $n_t = \nu n_{t-1} + (1-\nu) g_{t}^2$
		\STATE $m_t = \mu m_{t-1} + \lambda_t (1-\mu) \frac{ g_{t}}{\sqrt{{n}_t/c_n} + \epsilon} $
		%\STATE \#\#Following lines are the differences
		%\STATE $\mu_t = \beta_2 \mu_{t-1} + (1 - \beta_2) g_{t}$
		
		\STATE $\theta_{t+1} = (\theta_t -  m_t / c_m)\times(1-w_t) $
		\ENDFOR
	\end{algorithmic}
\end{algorithm}
\subsection{AmsProp}
\vspace{-2mm}
In Ref.~\cite{Reddi2018convergence}, the authors proposed \amsgrad\ as a variant of \adam\ that has a monotonically increasing $n_t$ to guarantee its convergence. This idea may be applied to \laprop\ similarly, which produces an algorithm that may be called {\it AmsProp}, as Algorithm~\ref{alg:amslaprop}. It should be noted that this algorithm subtly differs from the original \amsgrad. In practical cases, we have not observed the advantage of using such a variant, because a large $\nu$ can often do well enough by approximately producing a constant $n_t$ at convergence and therefore achieving a good performance. Nevertheless, AmsProp might be useful in special or complicated cases and we leave it for future research.
\begin{algorithm}[]
	\caption{AmsProp}
	\label{alg:amslaprop}
	\begin{algorithmic}
		\STATE {\bfseries Input:} $\theta_1 \in \mathbb{R}^d$, learning rate $\{\lambda_t\}_{t=1}^T$, decay parameters $0 \leq \mu < 1,\ 0 \leq \nu < 1 ,\ \epsilon \ll 1$, bias correction factors $0<c_n, c_m<1$. Set $m_{0} = 0$, $n_{0} = 0$, $\Tilde{n}_{0} = 0$.
		\FOR{$t=1$ {\bfseries to} $T$}
		%\STATE Draw a sample $s_t$ from $\mathbb{P}$.
		
		\STATE $g_{t} = \nabla_\theta \ell(\theta_{t})$
		\STATE $n_t = \nu n_{t-1} + (1-\nu) g_{t}^2$
		\STATE $\Tilde{n}_t = \max(\Tilde{n}_{t-1}, n_t)$
		\STATE $m_t = \mu m_{t-1} + \lambda_t(1-\mu) \frac{ g_{t}}{\sqrt{\Tilde{n}_t/c_n} + \epsilon} $
		%\STATE \#\#Following lines are the differences
		%\STATE $\mu_t = \beta_2 \mu_{t-1} + (1 - \beta_2) g_{t}$
		
		\STATE $\theta_{t+1} = \theta_t -  m_t / c_m $
		\ENDFOR
	\end{algorithmic}
\end{algorithm}
\subsection{Centered \laprop}
\vspace{-2mm}
Following the suggestion in Ref.~\cite{Tieleman2012_rmsprop}, we also propose a centered version of \laprop, which uses the estimation of the centered second moment rather than the non-centered second moment, which is a more aggressive strategy that would diverge in the presence of a constant gradient. The algorithm is given by Algorithm~\ref{alg:centered laprop}. As the estimation of the centered momentum is unstable at the initial steps, one may not do the update for the initial steps or one may use the original \laprop\ for the initial steps. 

Also, one can even combine the centered \laprop\ and \amsgrad, using the maximum of the centered second moment of the gradient. However, we have not been aware of any advantage of such a combination.

\begin{algorithm}[]
	\caption{Centered \laprop}
	\label{alg:centered laprop}
	\begin{algorithmic}
		\STATE {\bfseries Input:} $\theta_1 \in \mathbb{R}^d$, learning rate $\{\lambda_t\}_{t=1}^T$, decay parameters $0 \leq \mu < 1,\ 0 \leq \nu < 1 ,\ \epsilon \ll 1$, bias correction factors $0<c_n, c_m<1$, initial centered update step $t_\text{init}>1$. Set $m_{0} = 0$, $n_{0} = 0$, $\Bar{n}_{0} = 0$.
		\FOR{$t=1$ {\bfseries to} $T$}
		%\STATE Draw a sample $s_t$ from $\mathbb{P}$.
		
		\STATE $g_{t} = \nabla_\theta \ell(\theta_{t})$
		\STATE $n_t = \nu n_{t-1} + (1-\nu) g_{t}^2$
		\STATE $\Bar{n}_t = \nu \Bar{n}_{t-1} + (1-\nu) g_{t}$
		\IF{$t\geq t_\text{init}$}
		    \STATE $m_t = \mu m_{t-1} + \lambda_t(1-\mu) \frac{ g_{t}}{\sqrt{({n}_t-\Bar{n}_t^2)/c_n} + \epsilon} $
		\ELSE
		    \STATE $m_t = m_{t-1}$\quad or \quad $m_t=\mu m_{t-1} + \lambda_t(1-\mu) \frac{ g_{t}}{\sqrt{{n}_t/c_n} + \epsilon}$
		\ENDIF
		\STATE $\theta_{t+1} = \theta_t -  m_t / c_m $
		\ENDFOR
	\end{algorithmic}
\end{algorithm}

%\subsection{RaProp!}

\section{Additional Experiments and Experimental Details}
\vspace{-2mm}
We use the Pytorch library \cite{pytorch} for deep learning, and the codes that are used to produce our results are released on Github\footnote{\url{https://github.com/Z-T-WANG/LaProp-Optimizer}}.
\subsection{Achieving Better Generalization Performance}
\vspace{-2mm}
\begin{figure*}[tb]
\begin{subfigure}[b]{0.25\textwidth}
    \centering
    \includegraphics[trim=0 0 5 0, clip, width=1\textwidth]{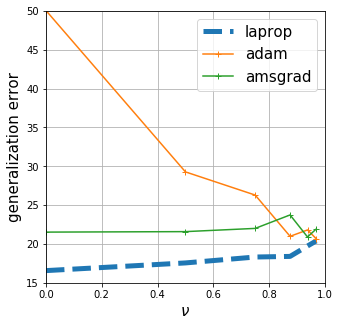}
    \vspace{-7mm}
    \caption{$r=0.0$, no label noise}
    \label{fig: imdb0}
\end{subfigure}
\hfill
\begin{subfigure}[b]{0.23\textwidth}
    \centering
    \includegraphics[trim=5 0 0 0, clip,width=1\textwidth]{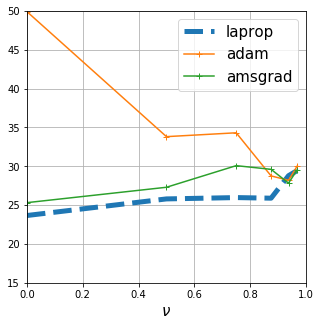}
    \vspace{-7mm}
    \caption{$r=0.1$}
    \label{fig: imdb1}
\end{subfigure}
\hfill
\begin{subfigure}[b]{0.23\textwidth}
    \centering
    \includegraphics[trim=5 0 0 0, clip, width=1\textwidth]{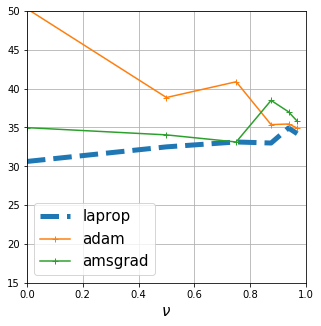}
    \vspace{-7mm}
    \caption{$r=0.2$}
    \label{fig: imdb2}
\end{subfigure}
\hfill
\begin{subfigure}[b]{0.23\textwidth}
    \centering
    \includegraphics[trim=5 0 0 0, clip, width=1\textwidth]{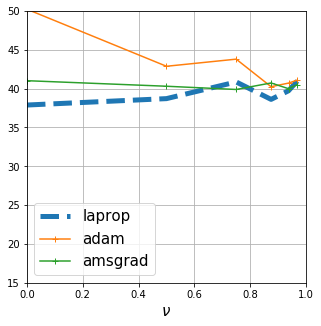}
    \vspace{-7mm}
    \caption{$r=0.3$}
    \label{fig: imdb3}
\end{subfigure}
\vspace{-1mm}
\caption{Generalization error on label-corrupted IMDB plotted against $\nu$, with flip probability $r$ of labels. For all the four corruption rates, the behaviors of \laprop\ are similar: a smaller $\nu$ gives a lower generalization error and better performance. For \adam, the performance gets worse for a smaller $\nu$ due to divergence and instability; for \amsgrad, the divergence is fixed and the learning is more stable than \adam, but its failure to decouple $\mu$ and $\nu$ seems to have a negative effect on performance when compared with \laprop.}\label{fig: imdb}
\vspace{-3mm}
\end{figure*}
While our analysis focuses on optimization, in this section we show how $\nu$ might be tuned to improve generalization. We consider the IMDB dataset, a binary classification task for which the goal is to identify the sentiment of the speakers, and it is a dataset where overfitting occurs quickly. We use LSTM \cite{bahdanau2014neural} with pretrained GloVe word embedding \cite{pennington2014glove}, and we also study how label noise affects the different optimizers in this task. The label of every data point is randomly flipped with probability $r\in\{0, 0.1, 0.2, 0.3\}$.

The training batch size is $128$ and the learning rate is set to $2\times 10^{-3}$ with $\mu=0.9$. All the models are trained for $15$ epochs. As shown in Figure~\ref{fig: imdb}, we find that (1) \laprop\ with $\nu=0$ always achieves the best generalization both with and without label noise, and especially at $r=0$, \laprop\ gives a $10\%$ accuracy improvement over the best possible hyperparameter setting of \adam\ and \amsgrad; (2) \adam\ and \amsgrad\ are not very stable w.r.t the changes in $\nu$, while \laprop\ responds to the changes in $\nu$ in a stable and predictable way, implying that \laprop's hyperparameter $\nu$ is easier to tune in practice.
\subsection{Optimization of the Noisy Rosenbrock Loss}
\begin{figure*}[tb]
	\hfill
	\begin{subfigure}[b]{0.28\textwidth}
		\centering
		\includegraphics[trim=0 0 0 0, clip, width=1\textwidth]{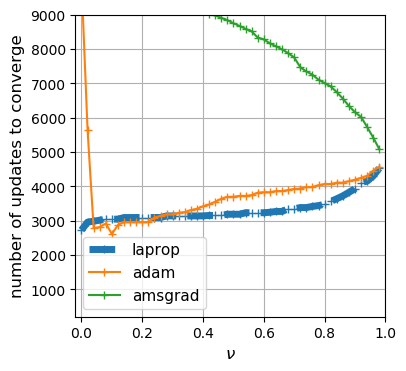}
		\vspace{-7mm}
		\caption{learning rate $4\times10^{-3}$}
	\end{subfigure}
	\hfill
	\begin{subfigure}[b]{0.265\textwidth}
		\centering
		\includegraphics[trim=20 0 0 0, clip,width=1\textwidth]{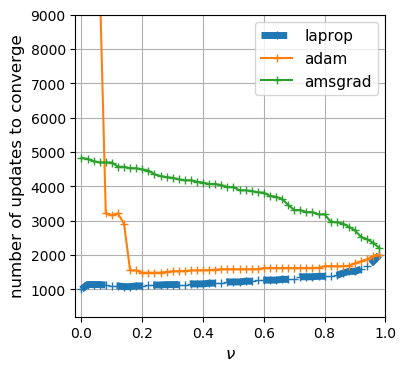}
		\vspace{-7mm}
		\caption{learning rate $1\times10^{-2}$}
	\end{subfigure}
	\hfill
	\begin{subfigure}[b]{0.265\textwidth}
		\centering
		\includegraphics[trim=20 0 0 0, clip, width=1\textwidth]{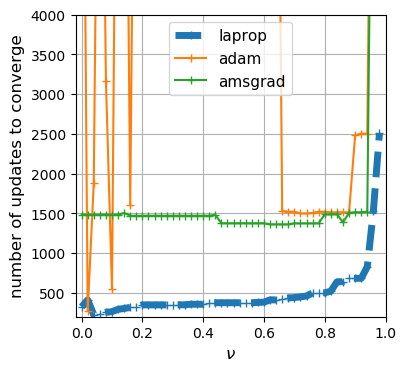}
		\vspace{-7mm}
		\caption{learning rate $4\times10^{-2}$}
	\end{subfigure}
	\hfill
	\vspace{-1mm}
	\caption{Time required for convergence on the noisy Rosenbrock task, with $\sigma=0.10$. A small learning rate makes \adam\ stabler but results in longer time for convergence, and in contrast, \laprop\ works stably across different magnitudes of learning rates.}\label{fig: rosen appendix}
	\vspace{-3mm}
\end{figure*}

\vspace{-2mm}
Most of the settings of the task are already described in the main text. To complete the details, the learning rate is set to be $10^{-2}$ in the main text, and we use $\mu=0.9$ and $\epsilon=10^{-8}$ for all the optimizers. Results for several different learning rates with $\sigma=0.1$ are shown in Fig.~\ref{fig: rosen appendix}.
\subsection{Neural Style Transfer}
\begin{figure*}[tb]
	\hfill
	\begin{subfigure}[b]{0.22\textwidth}
		\centering
		\includegraphics[trim=0 0 0 0, clip, width=0.9\textwidth]{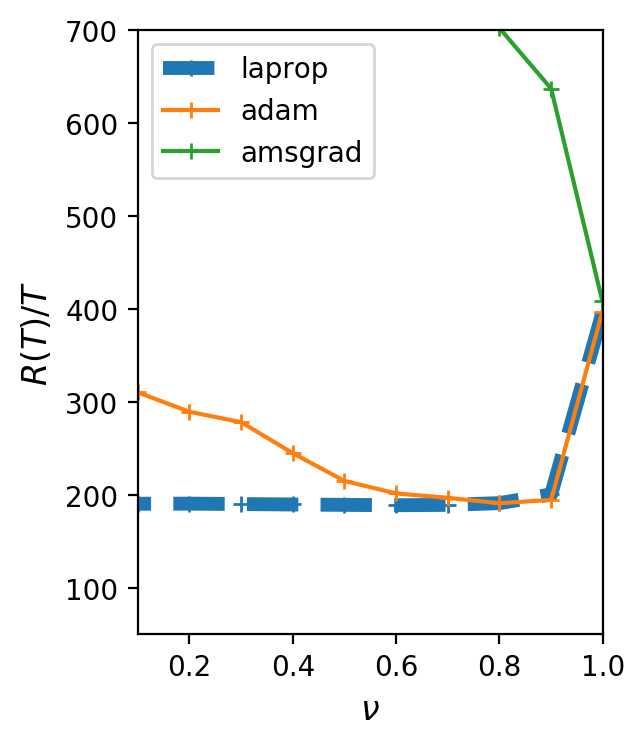}
		\vspace{-1mm}
		\caption{learning rate $1\times10^{-3}$}
	\end{subfigure}
	\hfill
	\begin{subfigure}[b]{0.203\textwidth}
		\centering
		\includegraphics[trim=20 0 0 0, clip,width=0.9\textwidth]{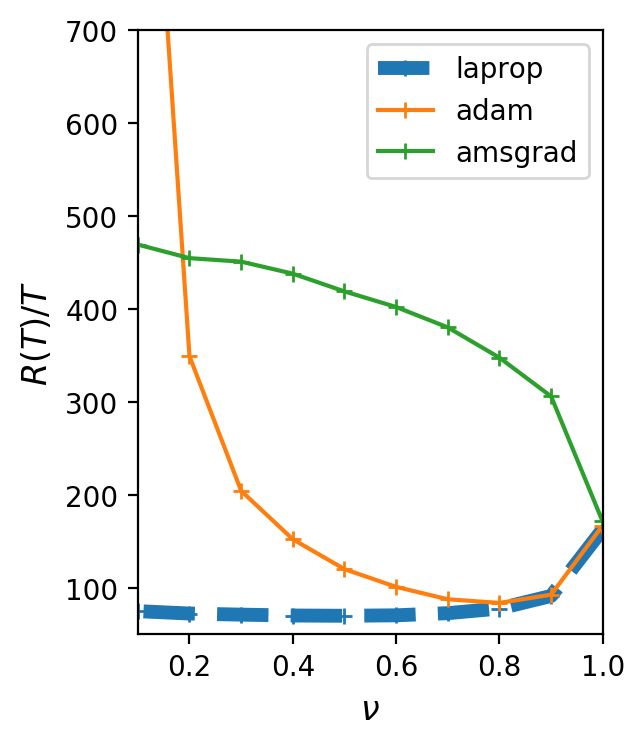}
		\vspace{-1mm}
		\caption{learning rate $1\times10^{-2}$}
	\end{subfigure}
	\hfill
	\begin{subfigure}[b]{0.203\textwidth}
		\centering
		\includegraphics[trim=20 0 0 0, clip, width=0.9\textwidth]{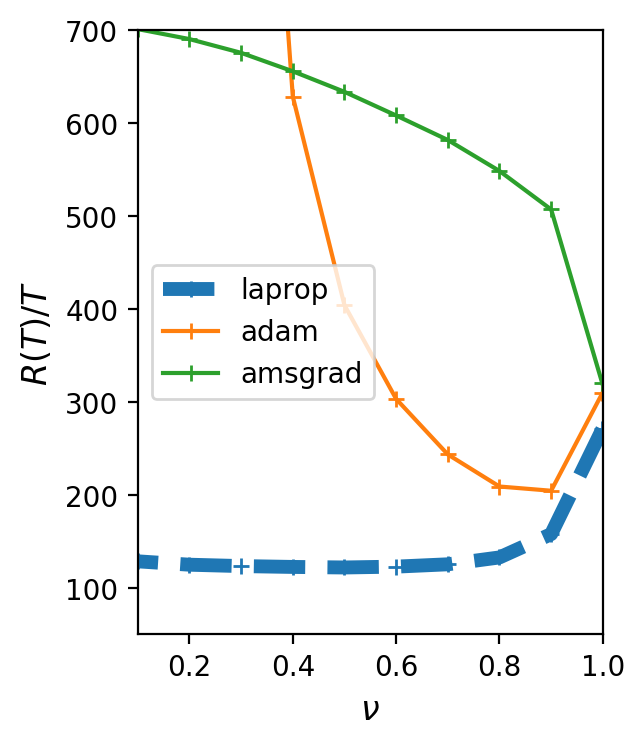}
		\vspace{-1mm}
		\caption{learning rate $2\times10^{-2}$}
	\end{subfigure}
	\hfill
	\vspace{-1mm}
	\caption{Neural style transfer with different optimizers and different learning rates. The average regret at $T=1000$ is plotted.}\label{fig:neural style appendix}
	\vspace{-3mm}
\end{figure*}

\vspace{-2mm}
We closely follow the prescription given in the Pytorch tutorial on neural transfer.\footnote{\url{https://pytorch.org/tutorials/advanced/neural_style_tutorial.html}} A pretrained VGG19 network is used to extract the features of a style image and a content image, and we optimize an input image so that its style follows the style image and its content follows the content image. The loss associated to content is calculated using the features extracted at the 4th convolutional layer of the network, and the loss associated to style is calculated as the average from the 1st to the 5th layer, using the normalized Gram matrices. All the settings follow the example given by Pytorch, except for the settings concerning the optimizers and except that a black empty image is used as our input image. A learning rate of $10^{-2}$ is used in the main text, and we use $\mu=0.9$ and $\epsilon=10^{-15}$, both for \adam\ and \laprop, and for \sgd\ we still keep using these parameters. Note that this optimization task is complex and usually L-BFGS is used as the optimizer. Results for other learning rates are shown in Fig.~\ref{fig:neural style appendix}.
\subsection{Training Extremely Deep Fully Connected Networks}
\vspace{-2mm}
The network has input dimension $784$, followed by $d$ layers all of which include $w=256$ hidden neurons, then followed by an output layer of dimension $10$, using the Kaiming initialization. The training batch size is $1024$. In our experiments, neither \laprop\ or \adam\ can learn with a learning rate larger than $1\times10^{-4}$, and as the learning rate decreases, \laprop\ begins to learn first, after which \adam\ also becomes able to learn. However, we find that the learning often struggles when the learning rate is too small, and a large learning rate produces better results for this problem provided that the learning can proceed. 
\subsection{Translation and Pretraining with Transformers}
\vspace{-2mm}
For the IWSLT task, we use the default settings provided by Ref.~\cite{ott2019fairseq} except for the learning rate schedule. As shown in Ref.~\cite{liu2019variance}, if we use a maximum learning rate of $3\times10^{-4}$, \adam\ will be trapped in a bad local minimum unless we use a warmup. Therefore, we use the $3\times10^{-4}$ learning rate to reproduce this result for \adam, and we also use it to test \laprop. After the warmup, or if no warmup is used, the learning rate linearly decreases and vanishes at the $60\times 10^3$-th update. Interestingly, we find that the result shown in the main text is highly learning-rate-dependent: if we use a maximum learning rate of $10^{-4}$ rather than $3\times10^{-4}$, \adam\ will not get trapped in a bad minimum and both \laprop\ and \adam\ can work well without a warmup schedule; if we use a maximum learning rate of $10^{-3}$, \laprop\ will not show a large difference from \adam\ as in the main text. Therefore, the advantage provided by \laprop\ does not remove the necessity of a warmup schedule, and the result in the main text is only for demonstrating the properties of \laprop. To plot the learning curves, we record the raw training loss of all update steps and apply a Gaussian smoothing with a standard deviation of 70 to smooth the data points.

For the RoBERTa task, the training batch size is 120 and the maximum learning rate is $10^{-4}$, and the learning rate linearly decreases and is expected to vanish at the $125\times 10^3$-th update, although we only carry out the initial $20\times 10^3$ updates. We use the momentum $\mu=0.9$ and the default RoBERTa setting $\epsilon=10^{-6}$ for \adam\ and the default \laprop\ setting $\epsilon=10^{-15}$ for \laprop. The maximum sequence length of data is kept 512 as default, and the trained model is Bert-base \cite{bert}, as provided by the Fairseq library \cite{ott2019fairseq}. We also enable the option of mask-whole-words in the Fairseq library to make the pretraining task more realistic. 

\begin{figure*}[tb]
\centering
\begin{subfigure}[]{0.495\textwidth}
\centering
    \includegraphics[width=\textwidth]{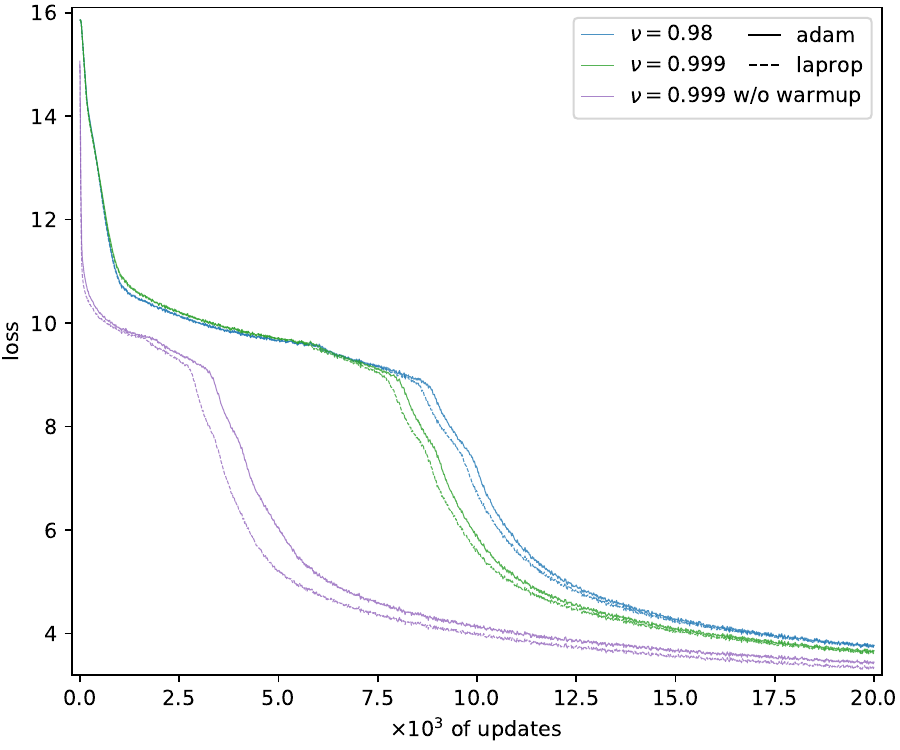}
    \caption{\label{fig: RobertaLarge}}
    \end{subfigure}
    \begin{subfigure}[]{0.495\textwidth}
    \centering
    \includegraphics[width=\textwidth]{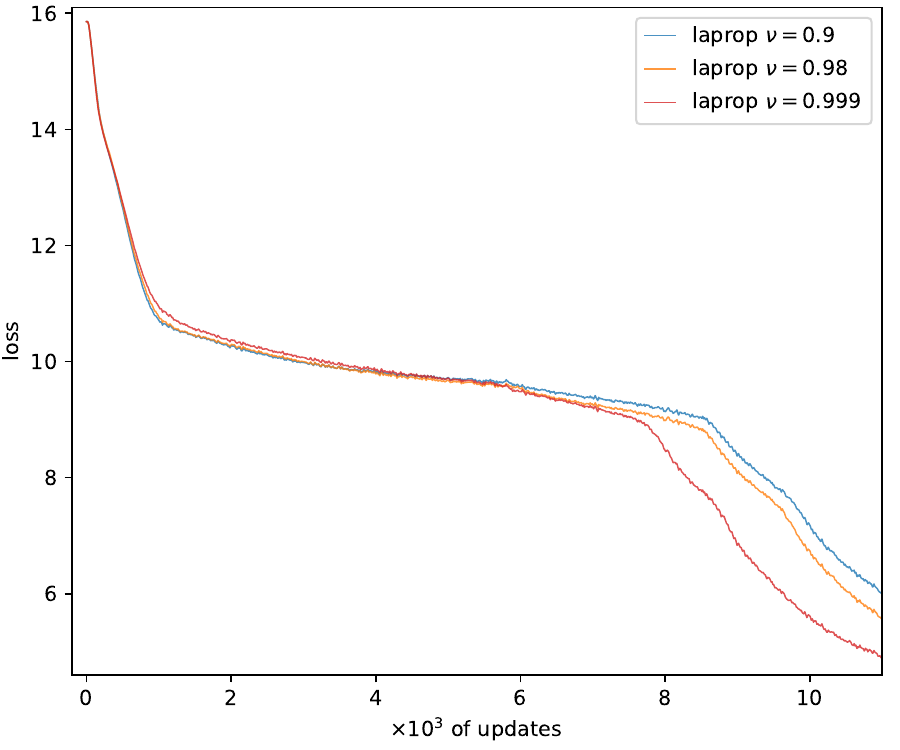}
    \caption{\label{fig: RobertaBeta2}}
    \end{subfigure}
    \caption{RoBERTa pretraining on the full English Wikipedia. See the text for details.}
    \label{fig: Roberta comparison}
\end{figure*}

We conjecture that the unnecessity of a warmup is due to the small learning rate and the relatively small variance of the training data, i.e.~only containing the English Wikipedia. In this case, we notice that a large $\nu$ may be used. We present the learning curves of different $\nu$ and using or not using warmup with more details in Fig.~\ref{fig: Roberta comparison}, training for the initial 20$\times10^3$ updates, which is actually less than one epoch. If one zooms in Fig.~\ref{fig: RobertaLarge}, it can also be observed that \laprop\ marginally outperforms \adam\ from an early stage. In Fig.~\ref{fig: RobertaBeta2}, it can be seen that a smaller $\nu$ accelerate the training at an early stage, while a larger $\nu$ converges better at a later stage. We have applied a Gaussian smoothing with a standard deviation $\sigma=6$ to the nearby data points in the plots, which is the same as in the main text. The fluctuation of the loss is much smaller than the case of IWSLT.

The English Wikipedia dataset is prepared following the instructions in the GitHub repository of GluonNLP \cite{gluoncvnlp2019}.\footnote{\url{https://github.com/dmlc/gluon-nlp/issues/641}} We used the latest English Wikipedia dump file and cleaned the text,\footnote{using scripts on \url{https://github.com/eric-haibin-lin/text-proc}} and we encode and preprocess the text following the standard RoBERTa procedure.\footnote{described in \url{https://github.com/pytorch/fairseq/blob/master/examples/roberta/README.pretraining.md}}
\subsection{Details of Reinforcement Learning Experiments}
\begin{figure*}[tb]
\centering
    \includegraphics[trim=9 12 12 5, clip, height=0.1425 \textwidth]{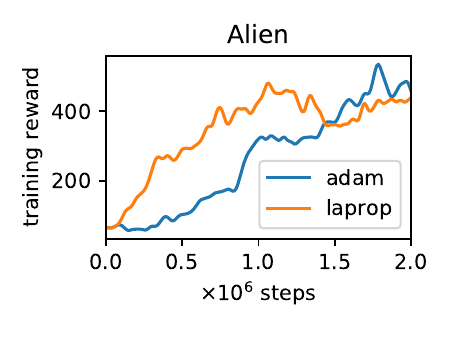}
    \includegraphics[trim=21 12 12 5, clip, height=0.1425 \textwidth]{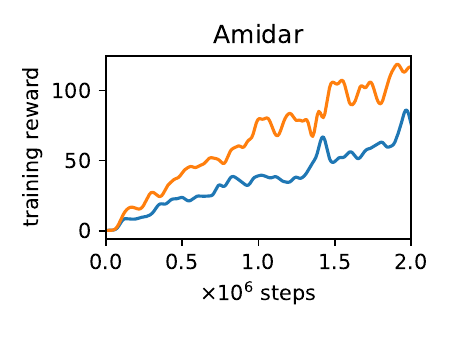}
    \includegraphics[trim=21 12 12 5, clip, height=0.1425 \textwidth]{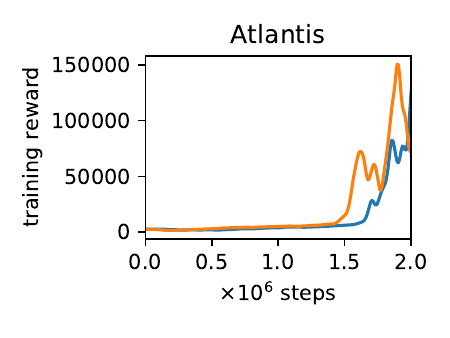}
    \includegraphics[trim=21 12 12 5, clip, height=0.1425 \textwidth]{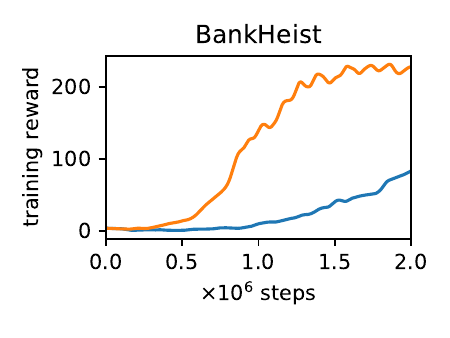}
    \includegraphics[trim=21 12 12 5, clip, height=0.1425 \textwidth]{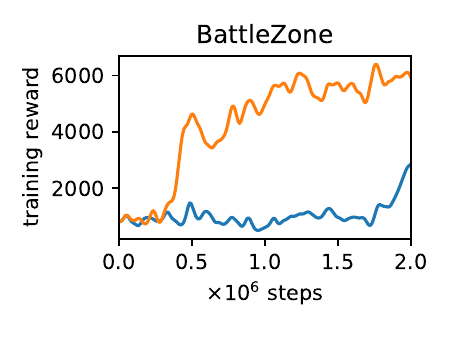}\\
    \includegraphics[trim=9 12 12 5, clip, height=0.1425 \textwidth]{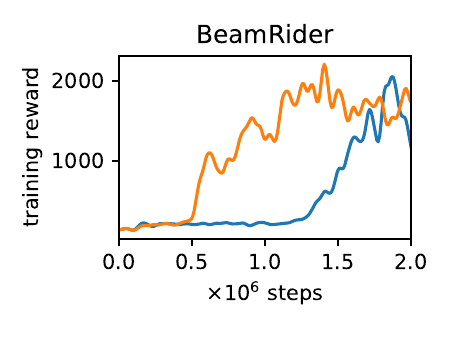}
    \includegraphics[trim=31 12 12 5, clip, height=0.1425 \textwidth]{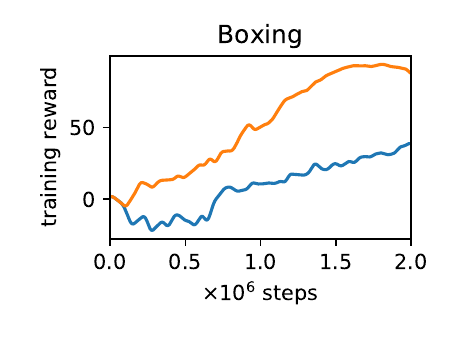}
    \includegraphics[trim=21 12 12 5, clip, height=0.1425 \textwidth]{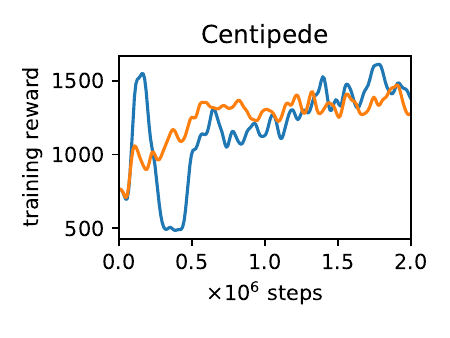}
    \includegraphics[trim=21 12 12 5, clip, height=0.1425 \textwidth]{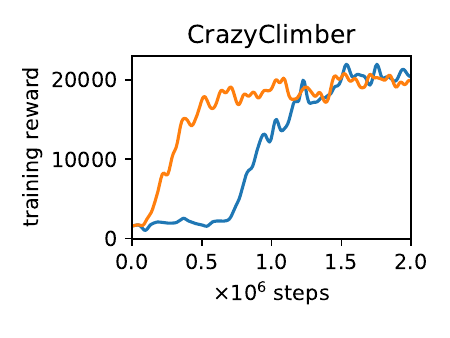}
    \includegraphics[trim=21 12 12 5, clip, height=0.1425 \textwidth]{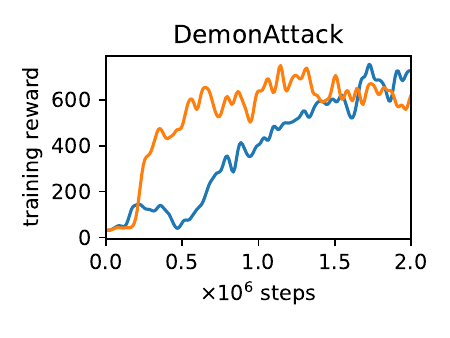}\\
    \includegraphics[trim=9 12 12 5, clip, height=0.1425 \textwidth]{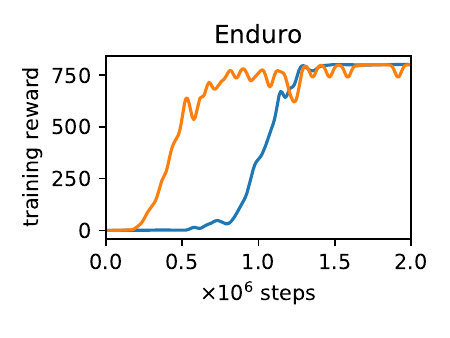}
    \includegraphics[trim=21 12 12 5, clip, height=0.1425 \textwidth]{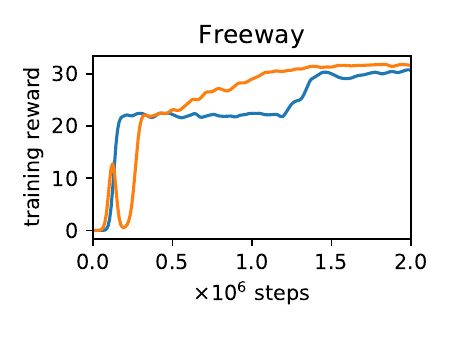}
    \includegraphics[trim=21 12 12 5, clip, height=0.1425 \textwidth]{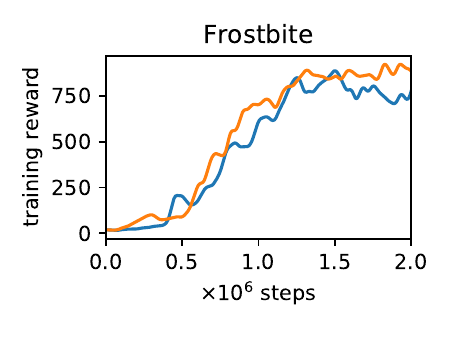}
    \includegraphics[trim=21 12 12 5, clip, height=0.1425 \textwidth]{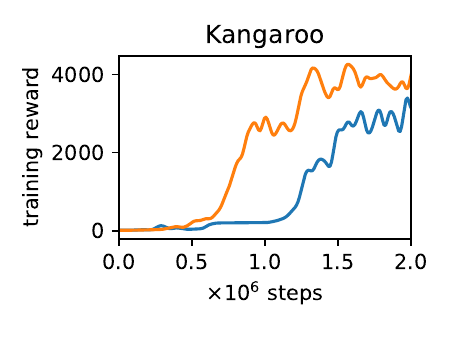}
    \includegraphics[trim=21 12 12 5, clip, height=0.1425 \textwidth]{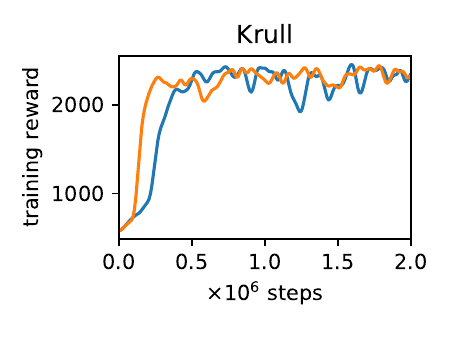}\\
    \includegraphics[trim=9 12 12 5, clip, height=0.1425 \textwidth]{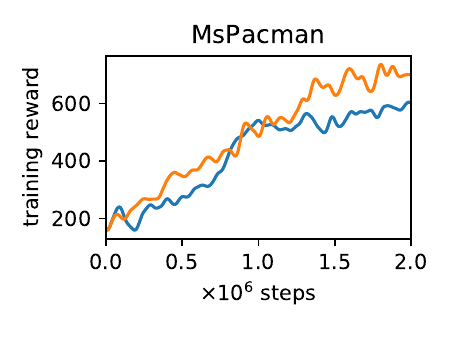}
    \includegraphics[trim=21 12 12 5, clip, height=0.1425 \textwidth]{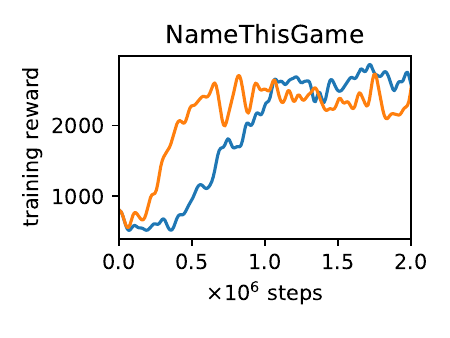}
    \includegraphics[trim=21 12 12 5, clip, height=0.1425 \textwidth]{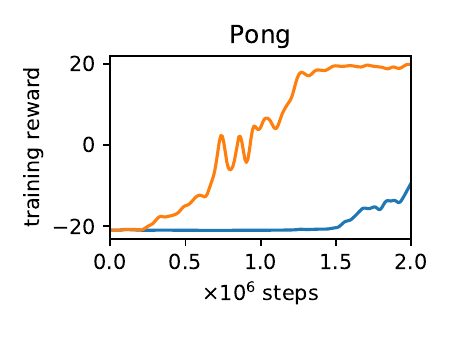}
    \includegraphics[trim=21 12 12 5, clip, height=0.1425 \textwidth]{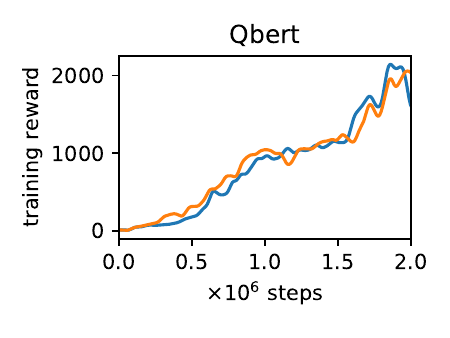}
    \includegraphics[trim=21 12 12 5, clip, height=0.1425 \textwidth]{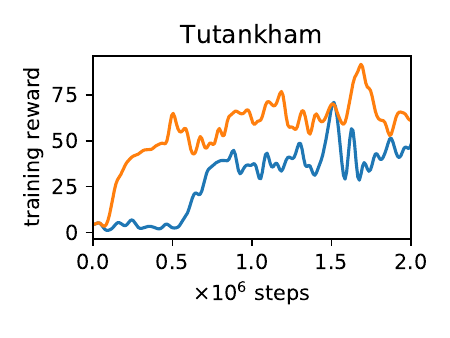}
    \vspace{-2mm}
    \caption{\label{fig: atari curves}Curves of training performance on 20 Atari2600 games, where the blue curves show the results of \adam\ and the orange curves show the results of \laprop. The training starts at the $8\times10^4$-th step, and per $10^4$ steps we record the average reward for a life in the game as the training performance here. Gaussian smoothing with $\sigma=2$ is applied to the curves.}
\end{figure*}

\vspace{-2mm}
The curves of training performance of the 20 Atari2600 games in our experiments are given in Fig.~\ref{fig: atari curves}. The random seeds for \laprop\ and \adam\ are always the same, so that the randomness in initialization and in the game process are removed. We can see that \laprop\ makes progress earlier than \adam.

Compared with Rainbow DQN \cite{rainbow}, we change the update period of target networks to 10000 steps (i.e., 40000 frames), and we avoid stop or loop of the games by forcing the agent to take random actions to lose a life if 10000 steps have passed in an episode. We also use a combined $\epsilon$-greedy and noisy network strategy, and the memory replay buffer only contains 0.5 million transitions and the buffer adopts a random replace strategy when it becomes full. Concerning the noisy network strategy, the trained network and the target network always use the same noise on training data to enhance consistency in their evaluations, which may have affected the performance marginally. Specifically for Atari2600 game {\it Breakout}, we do not use multi-step learning or the duel network structure, and the minimum value of $\epsilon$ in the greedy-$\epsilon$ strategy is reduced from 0.01 to 0.005.

Other parameter settings follow Ref.~\cite{rainbow}, except for the $\epsilon$ of \laprop, which is still the \laprop\ default $\epsilon=10^{-15}$. Therefore, $\epsilon$ of \adam\ is actually larger than that of \laprop. It is worth mentioning that the training batch size is 32, which is quite small, and that we have $\mu=0.9$ and $\nu=0.999$, with a learning rate of $6.25\times10^{-5}$.

\subsection{Image Classification on CIFAR10}
\begin{figure*}[tb!]
\centering
    \includegraphics[width=0.49 \textwidth]{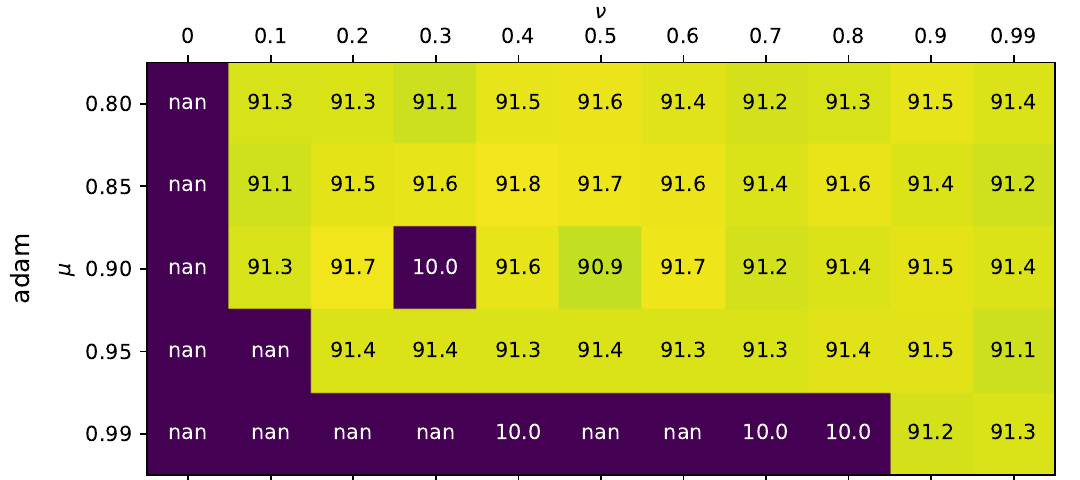}
    \includegraphics[width=0.49 \textwidth]{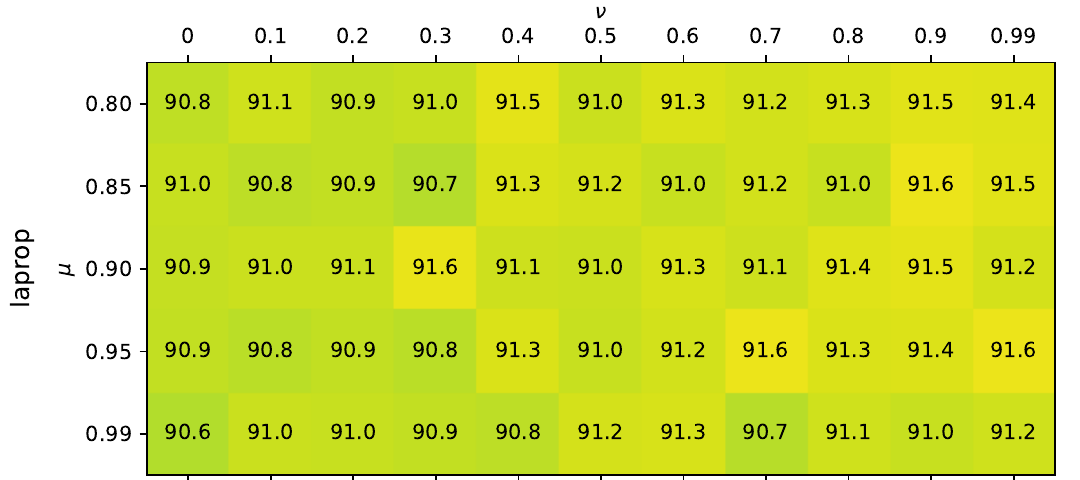}
    \caption{Test accuracy (\%) of Resnet-20 trained on CIFAR10, corresponding to the grid search models in Fig.~\ref{fig: CIFAR curves1}. {\it NAN} or {\it nan} is an abbreviation for {\it not a number}, which means that the machine encounters a numerical problem such as numerical overflow, and that it cannot obtain a number.
    \label{fig: CIFAR test accuracy large}}
    \vspace{-2mm}
\end{figure*}

\begin{figure*}[tb!]
\centering
\begin{subfigure}[]{\textwidth}
\centering
    \includegraphics[width=0.49 \textwidth]{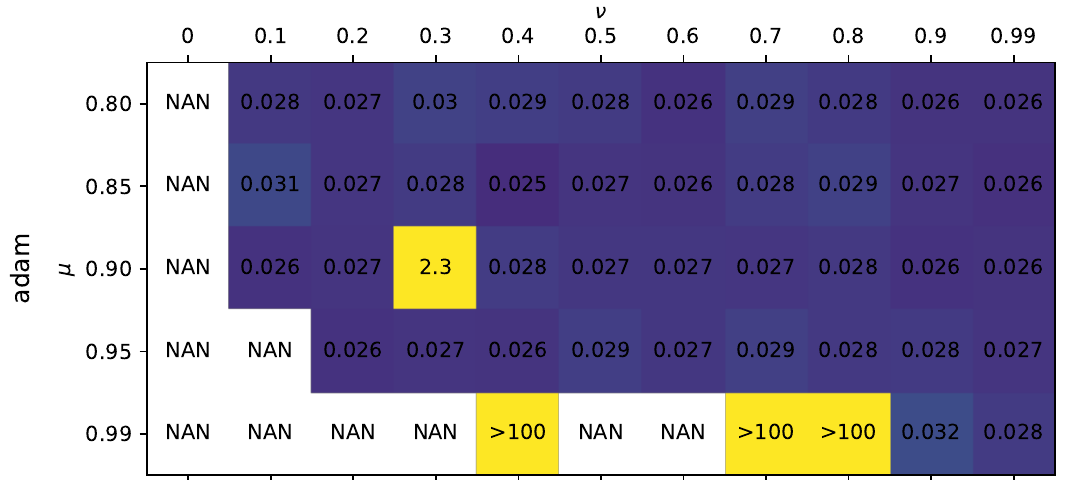}
    \includegraphics[width=0.49 \textwidth]{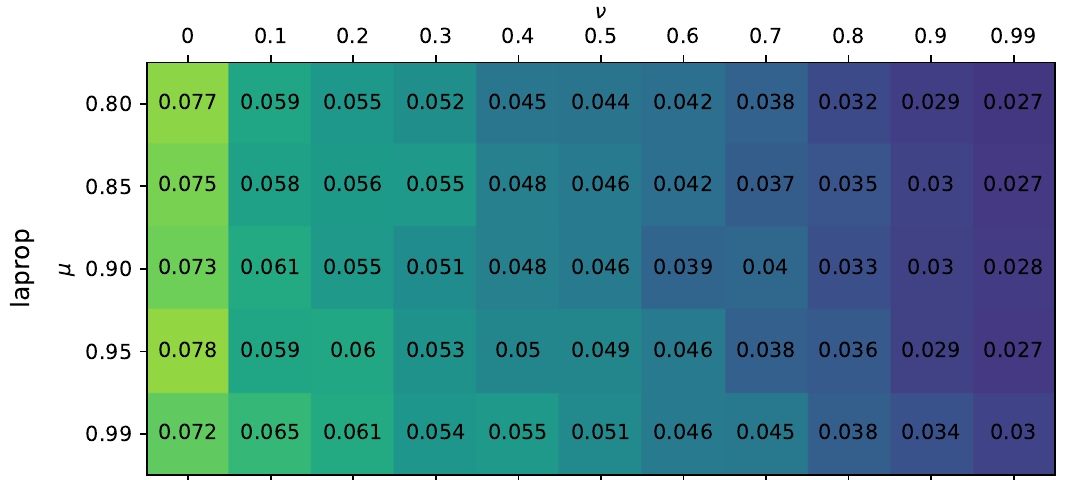}
    \vspace{-1mm}
    \caption{The training loss on CIFAR10 of \adam\ and \laprop \label{fig: CIFAR train loss table}}
\end{subfigure}
\begin{subfigure}[]{\textwidth}
\centering
    \includegraphics[width=0.49 \textwidth]{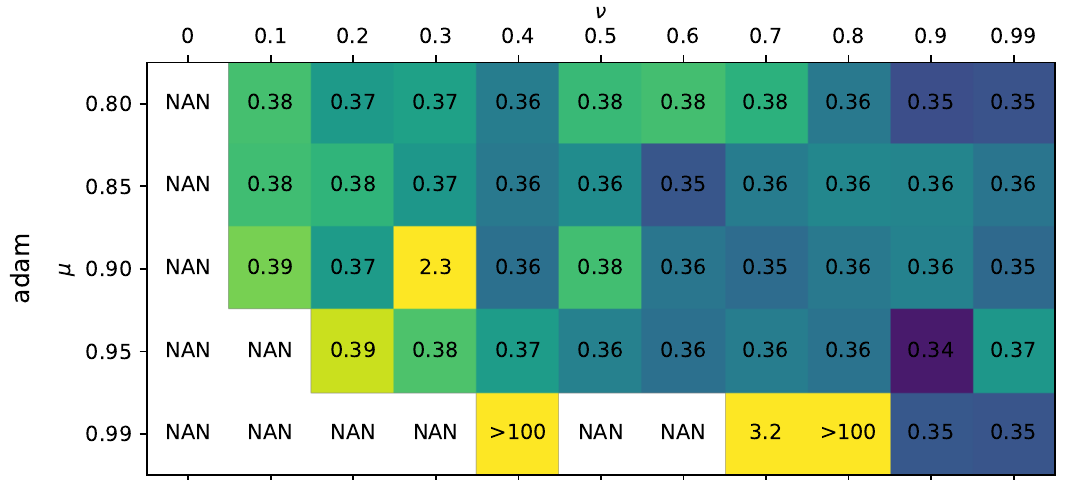}
    \includegraphics[width=0.49 \textwidth]{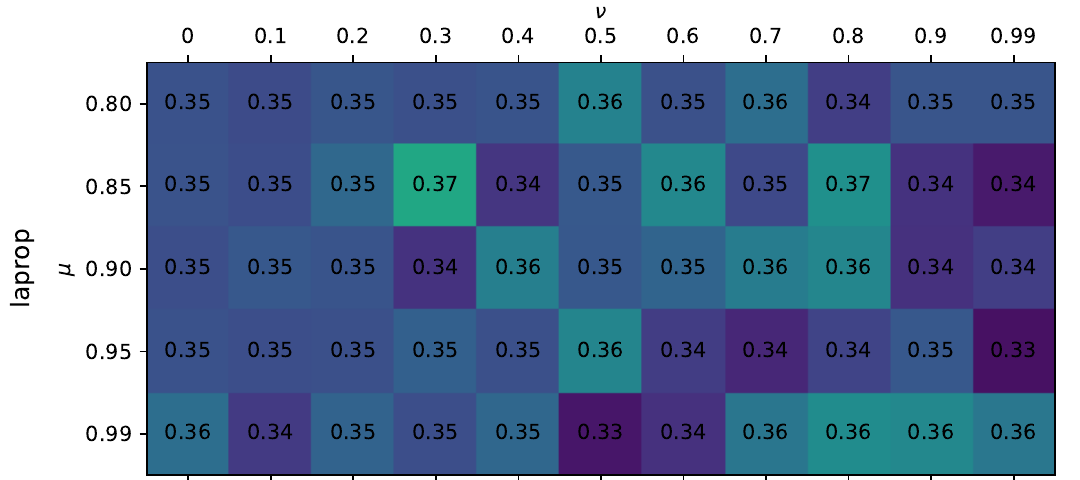}
    \vspace{-1mm}
    \caption{The test loss on CIFAR10 of \adam\ and \laprop \label{fig: CIFAR test loss table}}
\end{subfigure}
\vspace{-1mm}
    \caption{A summary of the final training and test loss in Fig.~\ref{fig: CIFAR curves1}. From the trend, we can clearly see that \adam\ tends to overfit in terms of loss when it is closer to divergence, while \laprop\ is not clearly affected. We see that if \adam\ does not diverge, \adam\ always reaches a low training loss irrespective of $\mu$ and $\nu$, while the training loss of \laprop\ is clearly dependent on $\mu$ and $\nu$. However, the test loss of \laprop\ is almost unaffected, but the test loss of \adam\ increases when it is closer to divergence, as shown for $\mu\le0.95$ and $0.1\le\nu\le0.4$, where the loss is higher than that of \laprop. This is an example where \laprop\ generalizes better than \adam. Surprisingly, we find that the higher test loss of \adam\ does not necessarily result in a worse test accuracy, which we think is probably a phenomenon specific to this task. \label{fig: CIFAR loss grid search}}
    \vspace{-2mm}
\end{figure*}

\vspace{-2mm}
We train deep residual networks on the CIFAR10 image classification task using \laprop\ and \adam\ \cite{he2016deep,cifar10}, implementing weight decay following the suggestions in Ref.~\cite{loshchilov2017fixing}. The network architecture is the standard Resnet-20,\footnote{We use the implementation in \url{https://github.com/bearpaw/pytorch-classification}} and we use the same hyperparameters for \adam\ and \laprop\ except for $\epsilon$ that is set to their default, and we perform a grid search on $\mu$ and $\nu$. The results on test accuracy are shown in Fig.~\ref{fig: CIFAR test accuracy large}. We find that \laprop\ and \adam\ perform comparably around the region of common hyperparameter choices, and when $\nu$ gets smaller \adam\ occasionally diverges, and if $\mu$ is much larger than $\nu$ then \adam\ diverges, while \laprop\ is not affected. We believe that the residual connections have made the model stable and more robust against divergence in this task. Interestingly, when we only look at the loss, we find that \adam\ tends to overfit when $\nu$ is small while \laprop\ does not, as shown in Fig.~\ref{fig: CIFAR loss grid search} and discussed in the caption. We have yet to find an explanation for this phenomenon.

\begin{figure*}[tb!]
\centering
    \includegraphics[width=0.1847 \textwidth]{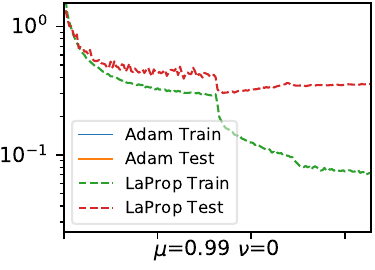}
    \includegraphics[width=0.155 \textwidth]{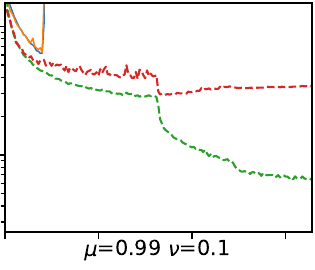}
    \includegraphics[width=0.155 \textwidth]{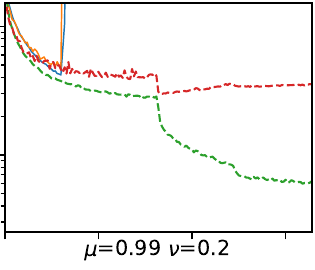}
    \includegraphics[width=0.155 \textwidth]{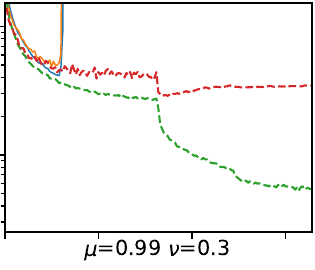}
    \includegraphics[width=0.155 \textwidth]{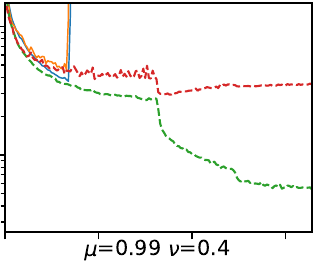}
    \includegraphics[width=0.155 \textwidth]{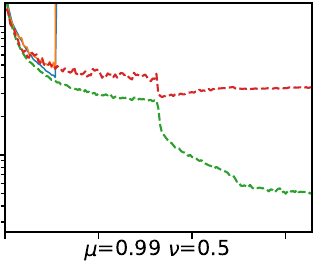}\\
    \includegraphics[width=0.185 \textwidth]{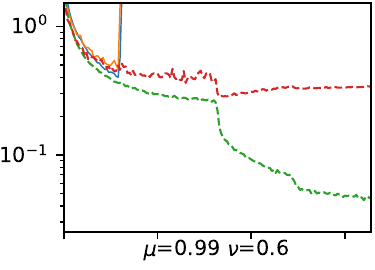}
    \includegraphics[width=0.155 \textwidth]{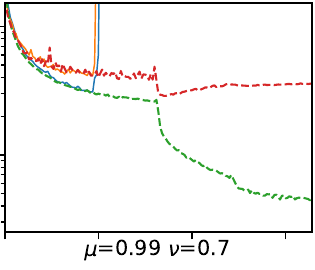}
    \includegraphics[width=0.155 \textwidth]{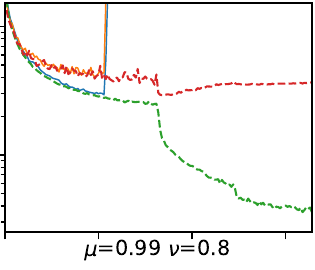}
    \includegraphics[width=0.155 \textwidth]{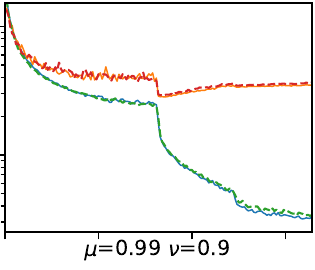}
    \includegraphics[width=0.155 \textwidth]{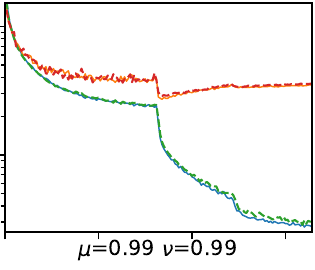}\\
    \vspace{3mm}
    \includegraphics[width=0.1847 \textwidth]{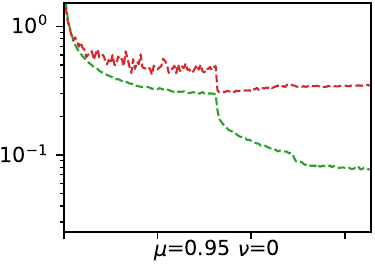}
    \includegraphics[width=0.155 \textwidth]{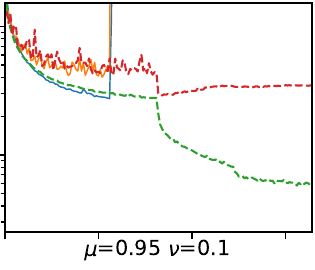}
    \includegraphics[width=0.155 \textwidth]{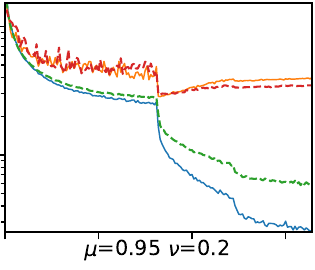}
    \includegraphics[width=0.155 \textwidth]{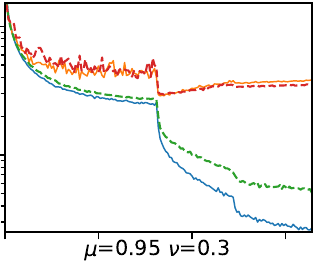}
    \includegraphics[width=0.155 \textwidth]{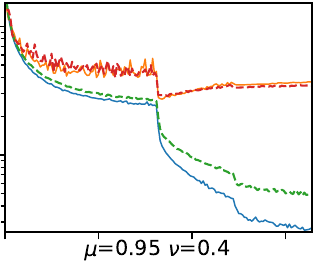}
    \includegraphics[width=0.155 \textwidth]{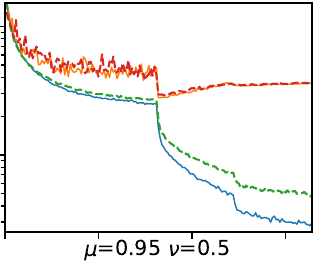}\\
    \includegraphics[width=0.185 \textwidth]{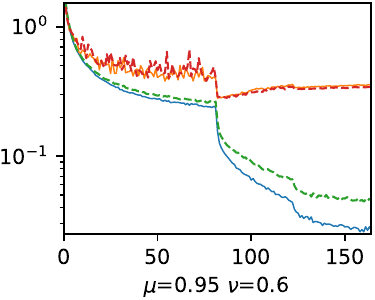}
    \includegraphics[width=0.155 \textwidth]{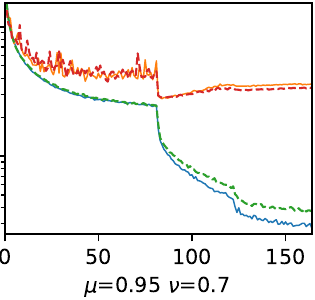}
    \includegraphics[width=0.155 \textwidth]{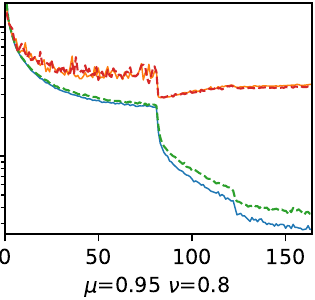}
    \includegraphics[width=0.155 \textwidth]{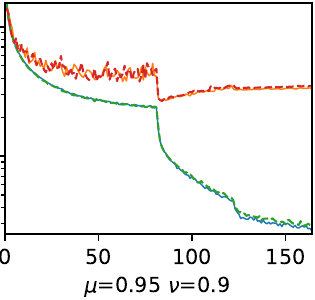}
    \includegraphics[width=0.155 \textwidth]{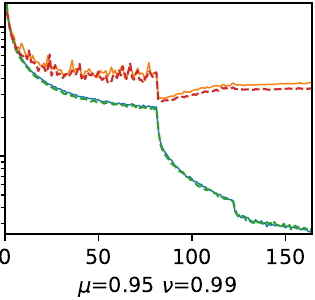}\\
    \vspace{3mm}
        \includegraphics[width=0.1847 \textwidth]{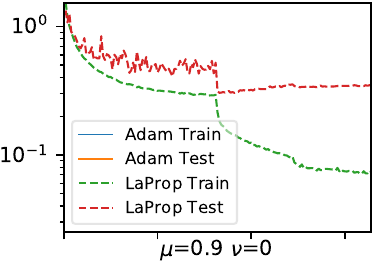}
    \includegraphics[width=0.155 \textwidth]{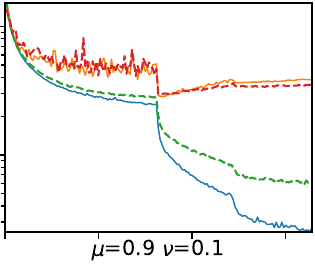}
    \includegraphics[width=0.155 \textwidth]{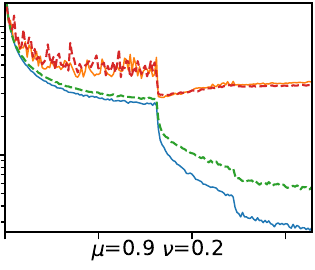}
    \includegraphics[width=0.155 \textwidth]{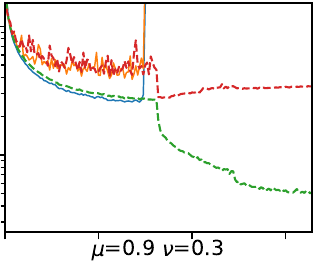}
    \includegraphics[width=0.155 \textwidth]{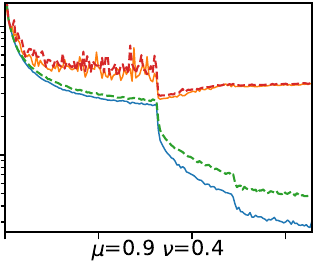}
    \includegraphics[width=0.155 \textwidth]{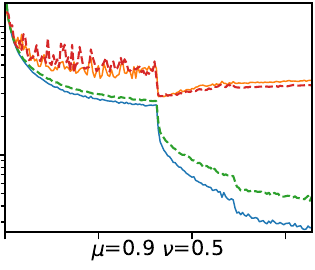}\\
    \includegraphics[width=0.185 \textwidth]{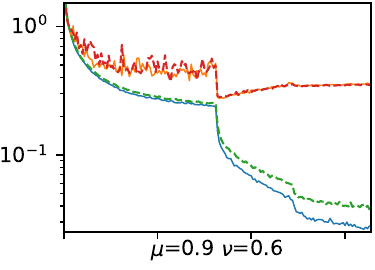}
    \includegraphics[width=0.155 \textwidth]{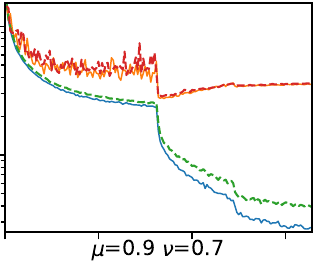}
    \includegraphics[width=0.155 \textwidth]{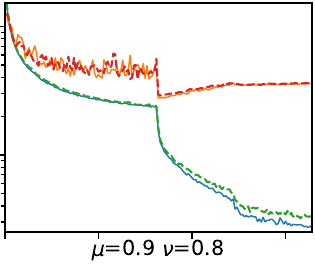}
    \includegraphics[width=0.155 \textwidth]{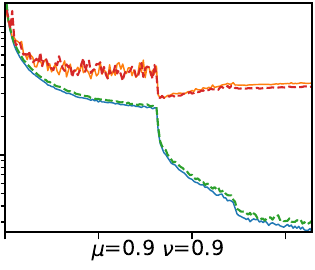}
    \includegraphics[width=0.155 \textwidth]{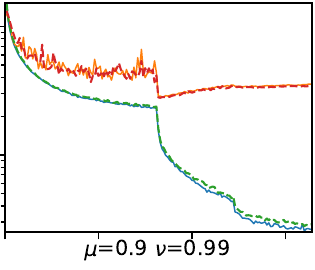}
    \vspace{-2mm}
    \caption{The learning curves of the Resnet-20 grid search on CIFAR10. The training loss and test loss are plotted for \adam\ and \laprop, and the meaning of the curves are shown in the legend in the first plot. If \adam\ diverges, its curves become absent in the plots. In the above figures, we see that when \adam\ diverges, a smaller $\nu$ causes the divergence to occur earlier. However, divergence sometimes occurs accidentally, such as the case of $\mu=0.9,\nu=0.3$. We see that the curves of \laprop\ and \adam\ resemble, except for that \adam\ often reaches a lower training loss. For the rest of the figures see Fig.~\ref{fig: CIFAR curves2}.}
    \label{fig: CIFAR curves1}
\end{figure*}

We use a learning rate of $10^{-3}$ and a weight decay of $10^{-4}$, and they are both reduced by a factor of 10 at epoch 80 and 120, and we train for 164 epochs and report the average of the last 5 epochs. Other settings follow Ref.~\cite{he2016deep}. The complete learning curves of training and test loss are given in Fig.~\ref{fig: CIFAR curves1}.

\begin{figure*}[tb!]
\centering
    \includegraphics[width=0.1847 \textwidth]{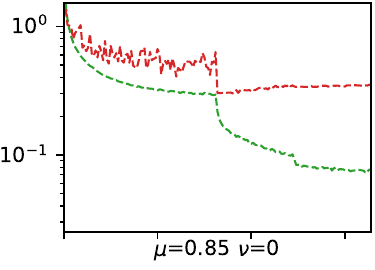}
    \includegraphics[width=0.155 \textwidth]{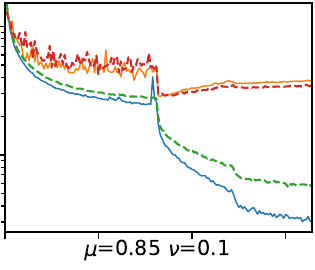}
    \includegraphics[width=0.155 \textwidth]{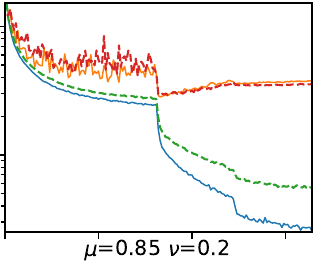}
    \includegraphics[width=0.155 \textwidth]{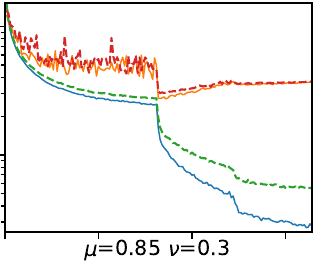}
    \includegraphics[width=0.155 \textwidth]{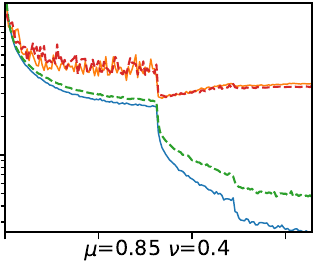}
    \includegraphics[width=0.155 \textwidth]{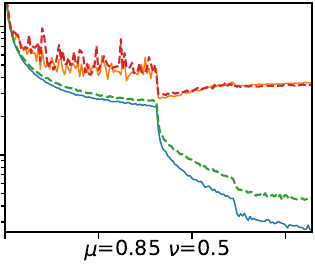}\\
    \includegraphics[width=0.185 \textwidth]{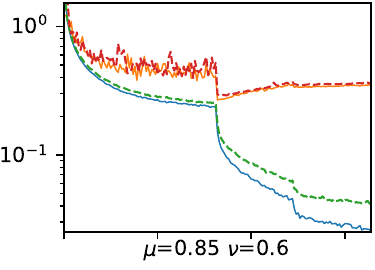}
    \includegraphics[width=0.155 \textwidth]{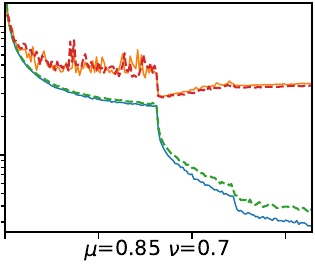}
    \includegraphics[width=0.155 \textwidth]{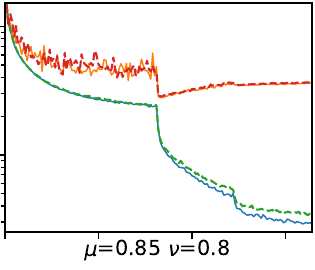}
    \includegraphics[width=0.155 \textwidth]{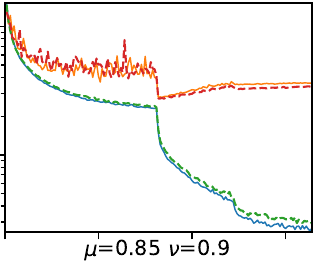}
    \includegraphics[width=0.155 \textwidth]{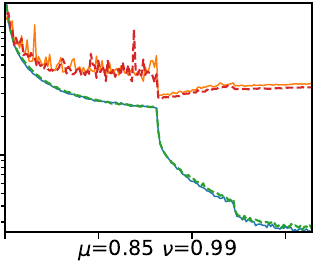}\\
    \vspace{3mm}
    \includegraphics[width=0.1847 \textwidth]{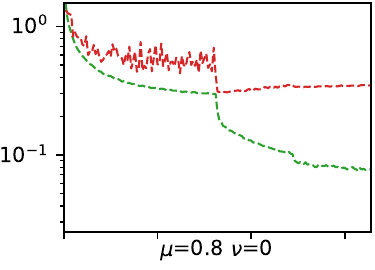}
    \includegraphics[width=0.155 \textwidth]{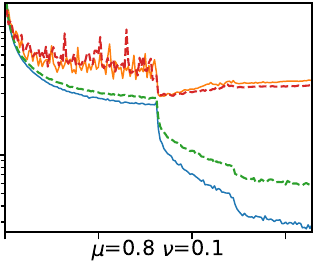}
    \includegraphics[width=0.155 \textwidth]{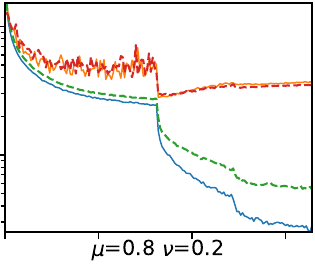}
    \includegraphics[width=0.155 \textwidth]{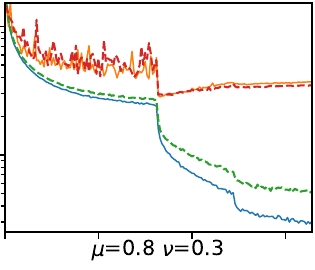}
    \includegraphics[width=0.155 \textwidth]{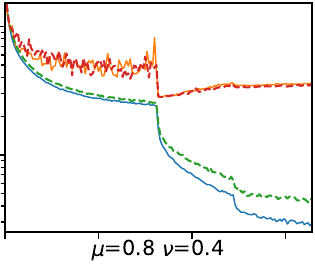}
    \includegraphics[width=0.155 \textwidth]{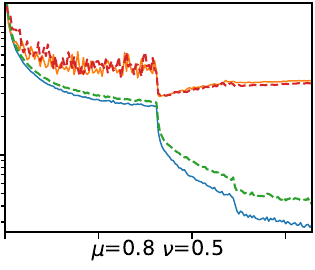}\\
    \includegraphics[width=0.1847 \textwidth]{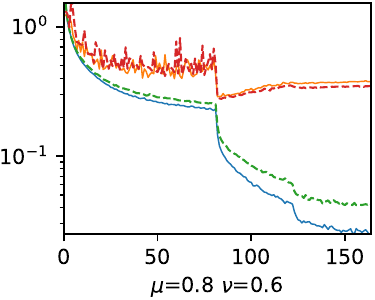}
    \includegraphics[width=0.155 \textwidth]{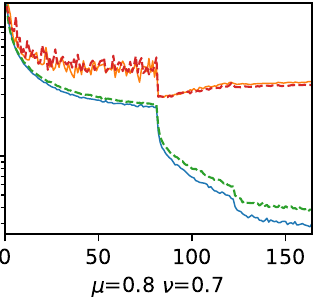}
    \includegraphics[width=0.155 \textwidth]{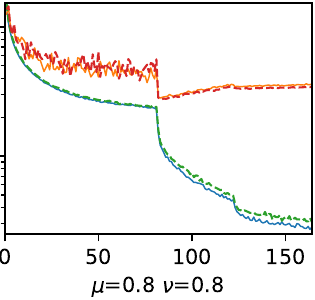}
    \includegraphics[width=0.155 \textwidth]{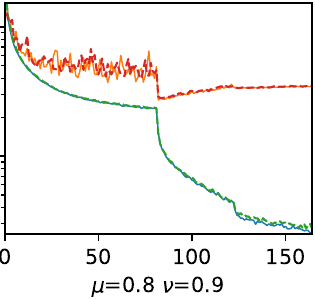}
    \includegraphics[width=0.155 \textwidth]{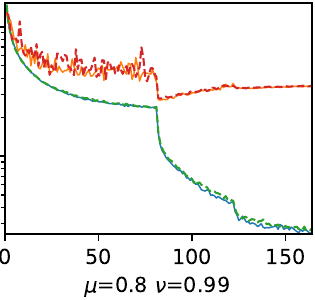}
    \vspace{-2mm}
    \caption{\label{fig: CIFAR curves2}}
    
\end{figure*}

\clearpage
%\bibliography{example_paper}
%\bibliographystyle{unsrt}

\end{document}